\documentclass[letterpaper]{article} % DO NOT CHANGE THIS
\usepackage{aaai2026}  % DO NOT CHANGE THIS
\usepackage{times}  % DO NOT CHANGE THIS
\usepackage{helvet}  % DO NOT CHANGE THIS
\usepackage{courier}  % DO NOT CHANGE THIS
\usepackage[hyphens]{url}  % DO NOT CHANGE THIS
\usepackage{graphicx} % DO NOT CHANGE THIS
\usepackage{natbib}  % DO NOT CHANGE THIS AND DO NOT ADD ANY OPTIONS TO IT
\usepackage{caption} % DO NOT CHANGE THIS AND DO NOT ADD ANY OPTIONS TO IT
\usepackage{newfloat}
\usepackage{listings}
\DeclareCaptionStyle{ruled}{labelfont=normalfont,labelsep=colon,strut=off} % DO NOT CHANGE THIS
\usepackage{array}

\usepackage{subcaption}
\usepackage{algorithm}
\usepackage{algpseudocode}
\usepackage{tabularx}
\usepackage{amsmath}
\usepackage{booktabs}
\usepackage{siunitx}
\usepackage{lipsum}
\usepackage{amsthm}
\usepackage{multicol}
\usepackage{placeins}
\usepackage{amsmath,amsfonts,bm}

\def\eqref#1{equation~\ref{#1}}
\def\1{\bm{1}}

\DeclareMathAlphabet{\mathsfit}{\encodingdefault}{\sfdefault}{m}{sl}
\SetMathAlphabet{\mathsfit}{bold}{\encodingdefault}{\sfdefault}{bx}{n}

\usepackage{multirow}
\usepackage{dsfont}

\newtheorem{Def}{Definition}
\newtheorem{Def2}[Def]{Definition}

\newtheorem{proposition}{Proposition}

\title{Enforcing Calibration in Multi-Output Probabilistic Regression with Pre-rank Regularization}

\author{
    Naomi Desobry \textsuperscript{\rm 1}\equalcontrib, 
    Elnura Zhalieva\textsuperscript{\rm 2}\equalcontrib, 
    Souhaib Ben Taieb\textsuperscript{\rm 1}\textsuperscript{\rm 2}
}
\affiliations{
    \textsuperscript{\rm 1}Department of Computer Science, University of Mons\\
    \textsuperscript{\rm 2}Department of Statistics and Data Science, Mohamed Bin Zayed University of Artificial Intelligence\\
    naomi.desobry@umons.ac.be, elnura.zhalieva@mbzuai.ac.ae, souhaib.bentaieb@mbzuai.ac.ae
    
}

\begin{document}
% The file aaai.sty is the style file for AAAI Press 
% proceedings, working notes, and technical reports.

\maketitle
\begin{abstract}
Probabilistic models must be well calibrated to support reliable decision-making. While calibration in single-output regression is well studied, defining and achieving multivariate calibration in multi-output regression remains considerably more challenging. The existing literature on multivariate calibration primarily focuses on diagnostic tools based on pre-rank functions, which are projections that reduce multivariate prediction-observation pairs to univariate summaries to detect specific types of miscalibration. In this work, we go beyond diagnostics and introduce a general regularization framework to enforce multivariate calibration during training for arbitrary pre-rank functions. This framework encompasses existing approaches such as highest density region calibration and copula calibration. Our method enforces calibration by penalizing deviations of the projected probability integral transforms (PITs) from the uniform distribution, and can be added as a regularization term to the loss function of any probabilistic predictor. Specifically, we propose a regularization loss that jointly enforces both marginal and multivariate pre-rank calibration. We also introduce a new PCA-based pre-rank that captures calibration along directions of maximal variance in the predictive distribution, while also enabling dimensionality reduction. Across 18 real-world multi-output regression datasets, we show that unregularized models are consistently miscalibrated, and that our methods significantly improve calibration across all pre-rank functions without sacrificing predictive accuracy.

\end{abstract}

\section{Introduction}

Probabilistic models output full predictive distributions rather than point estimates, enabling principled uncertainty-aware decision-making in domains such as meteorology, finance, and medical diagnosis \citep{Murphy01091984, ProbabilisticForecastsfromtheNationalDigitalForecastDatabase, 8734764, ONKAL1994350, Gulshan2016-vl, Guizilini2019-kw}. However, to be reliable, these predictions must be \emph{calibrated}, that is, their predicted probabilities must align with the observed frequencies.
% their stated probabilities must align with empirical frequencies.

%In single-output regression, where the target variable $Y \in \mathbb{R}$, calibration is well understood and can be rigorously assessed via tools like the Probability Integral Transform (PIT). Any deviations from calibration - referred to as \textit{miscalibration} - can be addressed either during training through regularization techniques derived from scoring rule decompositions (\citep{Wilks2018-jz,Wessel2025-je}), or post-hoc through recalibration methods such as isotonic regression (\citep{Kuleshov2018-qx}) or kernel-based adjustments(\citep{Dheur2023-it}). Throughout this work, we refer to the calibration in single-output regression as \emph{univariate calibration}.

In single-output regression, calibration is well understood and can be evaluated using tools such as the Probability Integral Transform (PIT). Deviations from perfect calibration, referred to as miscalibration, can be corrected either during training via regularization techniques based on scoring rule decompositions \citep{Wilks2018-jz,Wessel2025-je}, or post hoc using recalibration methods such as isotonic regression \citep{Kuleshov2018-qx} or kernel-based adjustments \citep{Dheur2023-it}. Throughout this work, we refer to calibration in single-output regression as univariate calibration.

%\emph{Multivariate calibration}, by contrast, refers to calibration in multi-output regression tasks where $Y \in \mathbb{R}^D$, and is more challenging to measure and evaluate. When predictions are multivariate, accurate specification of marginal distributions alone is not enough, the predictions must also capture the dependencies and joint structure across target dimensions. While several tools have been developed to assess specific aspects of multivariate calibration (\citep{chung2024samplingbased, Ziegel2013-al}), defining general-purpose, interpretable, and effective multivariate calibration criteria remains an open problem.

\emph{Multivariate calibration}, by contrast, concerns the calibration of a multivariate target and is considerably more difficult to evaluate and achieve. When producing probabilistic predictions for such targets, correctly specifying the marginal distributions is not sufficient; the predictions must also accurately capture the dependencies and joint structure across target dimensions. Although several tools have been proposed to assess specific aspects of multivariate calibration \citep{chung2024samplingbased, Ziegel2013-al}, defining general-purpose, interpretable, and effective calibration methods for the multivariate setting remains an open challenge.

%One approach to evaluating multivariate calibration leverages \emph{pre-rank functions}, scalar summaries of prediction-observation pairs that generalize univariate rank-based diagnostics to the multivariate setting (\citep{Allen2023-ig}). Each pre-rank describes a specific type of miscalibration, focusing for instance on marginal behavior or summary-statistics like location, scale, or dependence structure in the predictions. By projecting complex joint relationships in the multivariate predictions to interpretable scalar quantities, pre-ranks offer a flexible  tool to measure different aspects of probabilistic calibration.

One approach to evaluating multivariate calibration involves the use of pre-rank functions, which are scalar summaries of prediction-observation pairs that extend univariate rank-based diagnostics to the multivariate setting \citep{Allen2023-ig}. Each pre-rank targets a specific aspect of miscalibration, such as marginal calibration or discrepancies in summary statistics like location, scale, or dependence structure. By projecting complex multivariate predictions onto interpretable scalar quantities, pre-rank functions provide a flexible and general framework for assessing different dimensions of probabilistic calibration.

% toolkit for probing different facets of predictive reliability.

%In this work, we move beyond diagnostic tools and propose to directly enforce \emph{multivariate calibration} with respect to a collection of pre-rank functions during training. These pre-ranks capture application-specific desired properties of predictions, such as marginal accuracy, scale, and dependence structure, and are incorporated into the loss via a regularization term that penalizes miscalibration. We also introduce a novel pre-rank based on Principal Component Analysis (PCA), which projects prediction–observation pairs onto directions of maximal variance in the predictive distribution, capturing calibration along statistically meaningful directions. Empirically, our method consistently improves calibration across all pre-ranks without sacrificing predictive accuracy. With a simple modification, we show that our method can achieve both marginal and calibration with respect to pre-rank functions. Lastly, we demonstrate that our PCA-based pre-rank can also be used as a dimensionality reduction tool, enabling strong calibration performance even in lower-dimensional projection spaces. 

In this work, we go beyond diagnostic tools and propose a method to directly enforce \textit{multivariate calibration} by incorporating a regularization term into the training loss. This term penalizes miscalibration with respect to a collection of pre-rank functions. We further introduce a novel pre-rank based on Principal Component Analysis (PCA), which projects prediction-observation pairs onto directions of maximal variance in the predictive distribution, thereby capturing calibration along statistically meaningful directions. Additionally, we propose a regularization loss that jointly enforces both marginal and multivariate pre-rank calibration. When combined with the PCA-based pre-rank, our approach also enables dimensionality reduction and improves computational efficiency. Empirically, our method consistently improves calibration across all pre-rank functions without compromising predictive accuracy. We make the following main contributions:

\begin{itemize}
\item We conduct a large-scale empirical study on 18 real-world multi-output regression datasets to evaluate the probabilistic calibration of unregularized models across a diverse set of pre-rank functions.

\item We propose a general regularization framework that can be integrated into the training of any probabilistic predictor to enforce multivariate calibration with respect to user-specified pre-rank functions. Our approach also includes a joint regularization loss that enforces both marginal and multivariate calibration. When combined with our PCA-based pre-rank, the method detects calibration along the top principal components of the predictive covariance while also serving as a dimensionality reduction technique.

\item We validate our framework on 18 benchmark datasets and show that it consistently improves calibration across all pre-rank metrics without compromising predictive accuracy.

\end{itemize}

%The remainder of the paper is organized as follows. Section () provides background on probabilistic prediction, proper scoring rules, and calibration in both univariate and multivariate regression. Section () formalizes multivariate calibration using pre-rank functions. Section () provides the results of large-scale empirical study on probabilistic calibration for regression using multivariate data. Section () introduces our regularization-based methodology. Section () reviews related work. Lastly, Sections () and () detail the experimental setup and present results on benchmark datasets.

\section{Background}

We consider a multivariate regression setting where inputs \( X \in \mathcal{X} \subseteq \mathbb{R}^L \) and targets \( Y \in \mathcal{Y} \subseteq \mathbb{R}^D \) are jointly distributed. The target \( Y = (Y_1, \dots, Y_D) \) has dimension \( D \geq 1 \). Our goal is to estimate the true conditional distribution \( F_{Y|X} \) from a finite dataset \( \mathcal{D} = \{(X_i, Y_i)\}_{i=1}^N \).

%To this end, we define a \emph{probabilistic predictor} \( F_\theta : \mathcal{X} \rightarrow \mathcal{F} \), where \( \theta \) denotes model parameters and \( \mathcal{F} \subseteq \mathcal{P}(\mathbb{R}^D) \) is a class of admissible distributions over \( \mathbb{R}^D \) and \( \mathcal{P}(\mathbb{R}^D) \) denotes the space of probability distributions on \( \mathbb{R}^D \).

To this end, we define a \emph{probabilistic predictor} $F_\theta : \mathcal{X} \rightarrow \mathcal{F}$, where $\mathcal{F} \subseteq \mathcal{P}(\mathbb{R}^D)$ is a class of admissible probability distributions over $\mathbb{R}^D$, $\mathcal{P}(\mathbb{R}^D)$ denotes the space of all probability distributions on $\mathbb{R}^D$, and $\theta$ represents the model parameters. For any input $x \in \mathcal{X}$, the model outputs a predictive cumulative distribution function $\hat{F}_{Y|X=x} \in \mathcal{F}$, with corresponding density $\hat{f}_{Y|X=x}$. This distribution may be available in closed form (e.g., a multivariate Gaussian) or approximated via samples.

%For any input \( x \in \mathcal{X} \), the model returns a predictive cumulative distribution function \( \hat{F}_{Y|X=x} \in \mathcal{F} \) (with associated density \( \hat{f}_{Y|X=x} \)). This distribution can either be given in closed form (e.g., a multivariate Gaussian), or approximated via samples.

We learn the model parameters $\theta$ by minimizing a proper scoring rule over a training dataset, thereby encouraging the predictive distribution $F_\theta(x)$ to align with the true conditional distribution of $Y$ given $X = x$. A scoring rule $S : \mathcal{F} \times \mathcal{Y} \to \mathbb{R}$ assigns a numerical score to each predictive distribution $\hat{F} \in \mathcal{F}$ and observed outcome $y \in \mathcal{Y}$. It is called \textit{proper} if it is minimized in expectation when $\hat{F}$ equals the true distribution, and \textit{strictly proper} if the minimizer is unique. Two widely used examples are the negative log-likelihood (NLL), $\mathrm{NLL}(\hat{F}, y) = -\log \hat{f}(y)$, where $\hat{f}$ is the density of $\hat{F}$, and the energy score (ES),
$$
\mathrm{ES}(\hat{F}, y) = \mathbb{E}_{\hat{Y} \sim \hat{F}} \|\hat{Y} - y\| - \frac{1}{2} \mathbb{E}_{\hat{Y}, \hat{Y}' \sim \hat{F}} \|\hat{Y} - \hat{Y}'\|.
$$
This estimation strategy, known as \textit{optimum score estimation} \citep{Gneiting2007-cn}, allows flexible learning of predictive distributions. However, model misspecification and limited data may lead to biased or miscalibrated models, and the choice of scoring rule can also affect the accuracy, robustness, and calibration of the resulting model.

\noindent \textbf{Univariate calibration.} To better understand calibration in the multivariate setting, we briefly recall probabilistic calibration in the univariate setting. Let \( X \in \mathcal{X} \) and \( Y \in \mathbb{R} \) be random variables with conditional distribution \( F_{Y|X} \), and let \( \hat{F}_{Y|X} \) be a probabilistic predictor.
\begin{Def2}
\label{Def2}
\( \hat{F}_{Y|X} \) is said to be \emph{PIT-calibrated} if the probability integral transform (PIT),
\[
Z = \hat{F}_{Y|X}(Y),
\]
is uniformly distributed on \([0,1]\), that is,
\begin{equation}
F_Z(\alpha) = \alpha \quad \text{for all } \alpha \in [0,1].
\label{eq:pit_univariate}
\end{equation}
\end{Def2}
This property guarantees that the predicted distribution is statistically consistent with the observed outcomes. This condition holds if \( \hat{F}_{Y|X} \) matches the true conditional CDF. In practice, the deviation of the PIT distribution from uniformity can be quantified using the \emph{probabilistic calibration error} (PCE), defined as
\begin{equation}
\text{PCE}(F_\theta, \mathcal{D}) = \frac{1}{M} \sum_{j=1}^M \left| \alpha_j - \hat{F}_Z(\alpha_j) \right|, \label{eq:pce}
\end{equation}
where $\{\alpha_j\}_{j=1}^M$ is a grid of quantile levels such that $\alpha_j \in [0,1]$, and $\hat{F}_Z$ is the empirical CDF of the PIT values $Z_i = \hat{F}_{Y|X=X_i}(Y_i)$, given by $ \hat{F}_Z(\alpha) = \frac{1}{N} \sum_{i=1}^N \mathbf{1}(Z_i \leq \alpha)$.
Although this nonparametric estimator is effective for evaluation purposes, its non-differentiability prevents its direct application during training.

\paragraph{Regularization for univariate calibration.}

To improve the calibration of probabilistic models, regularization-based approaches add a calibration-specific penalty term to the training objective, explicitly encouraging the PIT values to follow a uniform distribution \citep{Wilks2018-jz,Dheur2023-it}. These methods aim to improve calibration, potentially at the expense of sharpness, with the trade-off controlled by a regularization hyperparameter.

Among such approaches, \citet{Dheur2023-it} introduced a differentiable \textbf{PCE-KDE} regularizer, which smooths the empirical CDF of the PIT values using a logistic kernel density estimator (KDE). Given PIT values $Z_i = \hat{F}_{Y|X}(Y_i)$, the smoothed CDF at a grid point $\alpha_j$ is defined as:
\begin{equation}
    \Phi_{\text{KDE}}(\alpha_j; \{Z_i\}_{i=1}^N) = \frac{1}{N} \sum_{i=1}^{N} \sigma\left( \tau(\alpha_j - Z_i) \right),
    \label{CDF_PIT}
\end{equation}
where $\sigma(z) = \frac{1}{1 + e^{-z}}$ is the sigmoid function, and $\tau > 0$ controls the smoothness of the approximation. The resulting regularization term is given by:
\begin{equation}
    \mathcal{R}_\text{PCE-KDE} = \frac{1}{M} \sum_{j=1}^M \left| \alpha_j - \Phi_{\text{KDE}}(\alpha_j; \{Z_i\}_{i=1}^N) \right|^p,
\label{PCE-KDE-uni}
\end{equation}
where $p \geq 1$ determines the penalty's shape. Minimizing this term encourages the PIT distribution to align with the uniform distribution, thereby promoting probabilistic calibration during training.

\section{Related Work}
\noindent \textbf{Univariate Calibration.} A variety of methods have been proposed for univariate calibration, including post-hoc recalibration techniques \citep{Kuleshov2018-qx, Kuleshov2021-qy, Song2019-rx} and regularization-based approaches \citep{Zhao2020-pa, Feldman2021-aa}. \citet{Dheur2024-lw} provide a unified perspective on these methods and introduce a training framework that integrates recalibration directly into the learning process. Tail-focused calibration has also gained attention, particularly through the use of weighted scoring rules and loss regularization \citep{Wessel2025-je}. More recently, hybrid strategies that combine refinement during training with post-hoc calibration have been shown to improve both sharpness and reliability \citep{berta2025rethinkingearlystoppingrefine}.

\noindent \textbf{Multivariate Calibration.} Extending calibration to multivariate outputs is substantially more challenging due to the need to capture joint dependencies among target dimensions. One approach, known as \textit{copula calibration}, evaluates the uniformity of the copula PIT, generalizing univariate rank histograms to the multivariate setting by assessing the joint CDF of the predicted distribution \citep{Ziegel2013-al}. While conceptually appealing, no practical method currently exists to enforce copula calibration during training. An alternative line of work introduces \textit{pre-rank functions}, which project multivariate forecast-observation pairs to scalar quantities before constructing rank histograms \citep{Allen2023-ig}. These functions enable diagnostic assessment of specific forms of miscalibration but do not provide a mechanism for enforcing calibration under a given pre-rank. More recently, \textit{HDR calibration} has been proposed to target high-density regions of the predictive distribution \citep{chung2024samplingbased}. It operates post-hoc by learning a mapping that resamples predictions to satisfy HDR calibration, followed by a correction step that updates the predictive model itself, though this step currently applies only to Gaussian outputs.

%\noindent \textbf{Multivariate Calibration.} Extending calibration to multivariate outputs is more complex due to joint dependencies. \textit{Copula calibration} assesses the uniformity of the copula PIT and generalizes rank histograms to the multivariate case \citep{Ziegel2013-al}, capturing the full dependency structure via the joint CDF. However, no known method currently exists to enforce this property in practice. Another line of work introduces the use of \textit{pre-rank functions}, which map multivariate forecast–observation pairs to scalar quantities prior to constructing a rank histogram \citep{Allen2023-ig}, yet do not suggest how to achieve calibration under a given pre-rank. \textit{HDR calibration} targets high-density regions\citep{chung2024samplingbased}, and enforces calibration by learning a transformation toward uniformity. A correction step is further proposed to adjust the predictive model itself, but is restricted to Gaussian outputs \citep{chung2024samplingbased}. 

\section{Multivariate calibration with pre-ranks}

A closer examination of the multivariate calibration methods discussed in the previous section reveals that they can be interpreted within a unified framework based on \textit{pre-rank functions}. These are univariate functionals $\rho : \mathcal{X} \times \mathbb{R}^D \to \mathbb{R}$ that map multivariate forecast-observation pairs to scalar values for calibration assessment. Each pre-rank highlights a specific structural aspect of the predictive distribution. Let \((X, Y) \sim F_{Y|X}\) and define
\[
T = \rho(X, Y) \quad \text{and} \quad \hat{T} = \rho(X, \hat{Y}),
\]
where \( \hat{Y} \sim \hat{F}_{Y|X} \) denotes a sample from the predictive distribution. 
\begin{Def}
\label{Def1}
$\hat{F}_{Y|X}$ is said to be \textbf{calibrated with respect to a pre-rank \(\rho\)} if \( \hat{F}_{T|X} \) is PIT-calibrated (see Definition~\ref{Def2})
\end{Def}

As shown in \citet{chung2024samplingbased}, if the predictive distribution matches the true conditional distribution, i.e., \( \hat{F}_{Y|X} = F_{Y|X} \), then calibration holds for any choice of pre-rank $\rho$. In Table \ref{tab:preranks}, we present several pre-rank functions previously introduced in the literature and considered in this work.

\begin{table}[!htbp]
\centering
\small
\begin{tabular}{l|l}
\hline
Pre-rank & Formula \\
\hline
Marginal &  $\rho_{\text{marg}}^d(x, y) = y_d$ \\
Location &  $\rho_{\text{loc}}(x, y) = \frac{1}{D} \sum_{d=1}^D y_d$ \\
Scale &  $\rho_{\text{scale}}(x, y) = \frac{1}{D} \sum_{d=1}^D (y_d - \bar{y})^2$ \\
Dependency & $\rho_{\text{dep}}(x, y; h) = -\frac{\gamma_y(h)}{s_y^2}$ \\
HDR & $\rho_{\text{hdr}}(x, y) = \hat{f}_{Y|X=x}(y)$ \\
Copula & $\rho_{\text{cop}}(x, y) = \hat{F}_{Y|X=x}(y)$ \\
\hline
\end{tabular}
\caption{Types of pre-rank functions considered in this work.}
\label{tab:preranks}
\end{table}
The marginal pre-rank assesses calibration along individual dimensions by extracting the \(d\)-th coordinate for each \(d \in \{1, \dots, D\}\). The location pre-rank averages across dimensions to evaluate global bias, while the scale pre-rank measures the overall spread. The dependency pre-rank captures structural dependencies via a normalized variogram. For \(h \in \{1, \dots, D-1\}\) it is defined as $\gamma_y(h) = \frac{1}{2(D - h)} \sum_{d=1}^{D - h} |y_d - y_{d+h}|^2$, where $s^2_y$ is a variance across dimensions and acts as a normalizer.
The HDR pre-rank~\citep{chung2024samplingbased} adopts a likelihood-based perspective by evaluating the predicted density at the observed outcome. The Copula pre-rank, on the other hand, evaluates the predicted CDF at the observation, capturing the structure of the joint predictive distribution.

% \noindent \textbf{Marginal pre-rank}: for any \( d \in \{1, \dots, D\} \), extracts the \(d\)-th coordinate to assess calibration dimension-wise:
%     \[
%     \rho_{\text{marg}}^d(x, y) = y_d.
%     \]

% \noindent \textbf{Location pre-rank}: averages across dimensions to check global bias:
%     \[
%     \rho_{\text{loc}}(x, y) = \frac{1}{D} \sum_{d=1}^D y_d.
%     \]
    
% \noindent \textbf{Scale pre-rank}: measures total spread across dimensions:
%     \[
%     \rho_{\text{scale}}(x, y) = \frac{1}{D} \sum_{d=1}^D (y_d - \bar{y})^2.
%     \]
    
% \noindent \textbf{Dependency pre-rank}: evaluates the dependency structure via a normalized variogram. For any \( h \in \{1, \dots, D-1\} \),
%     \[
%     \rho_{\text{dep}}(x, y; h) = -\frac{\gamma_y(h)}{s_y^2}, \]
%     \[
%     \text{where} \quad \gamma_y(h) = \frac{1}{2(D-h)} \sum_{d=1}^{D-h} |y_d - y_{d+h}|^2.
%     \]

% \noindent \textbf{HDR pre-rank} \citep{chung2024samplingbased}: takes a likelihood-based view by evaluating the predicted density at the observed outcome:
%     \[
%     \rho_{\text{hdr}}(x, y) = \hat{f}_{Y|X=x}(y).
%     \]

% \noindent \textbf{Copula pre-rank} \citep{Ziegel2013-al}: evaluates the joint CDF value under the model,
%     \[
%     \rho_{\text{cop}}(x, y) = \hat{F}_{Y|X=x}(y),
%     \]
% which captures the dependence structure of the joint distribution.

These techniques offer complementary insights into the quality of probabilistic predictions by evaluating how well the model captures structural or distributional aspects of the output. However, we emphasize that these pre-rank functions are primarily diagnostic in nature. The existing literature does not offer a principled way to incorporate them into the training process to enforce multivariate calibration. This is precisely the gap our work addresses.
% Before introducing our approach, the next section demonstrates that standard probabilistic models, even when trained using strictly proper scoring rules, often fail to achieve calibration across the various pre-ranks introduced earlier. This highlights the need for dedicated recalibration methods to ensure that predictive distributions are not only sharp but also well-calibrated.

% 0.33\linewidth
\begin{figure*}[!htbp]
    \centering
    \begin{subfigure}[b]{0.33\linewidth}\includegraphics[width=\linewidth]{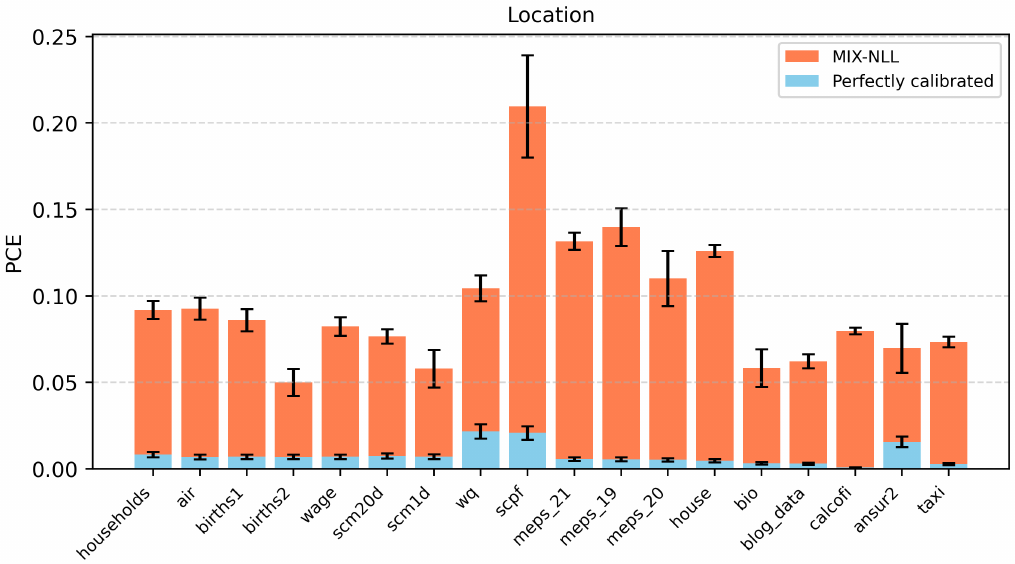}
        \caption{Location}
        \label{subfig:location}
    \end{subfigure}
    \begin{subfigure}[b]{0.33\linewidth}
        \includegraphics[width=\linewidth]{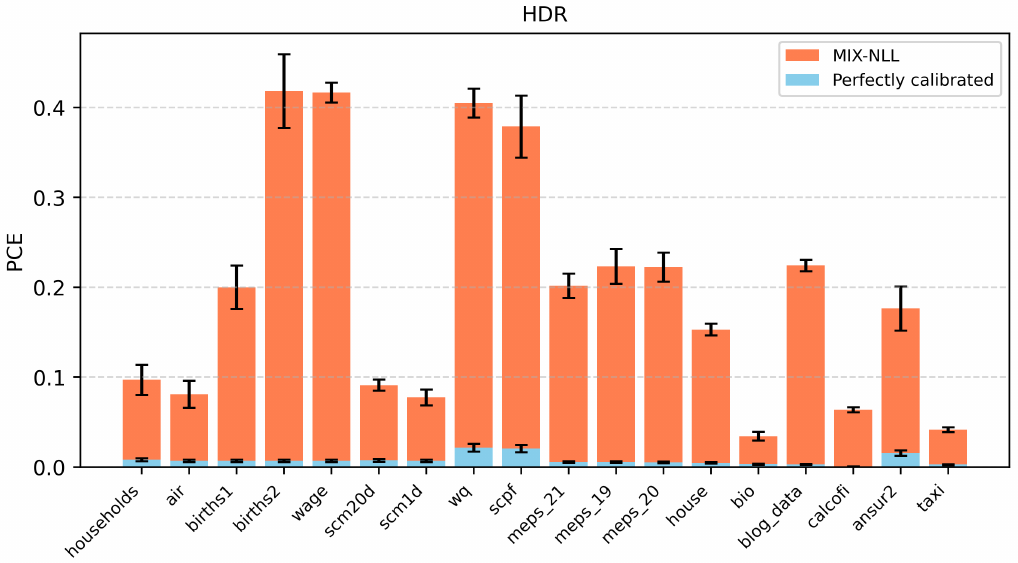}
        \caption{HDR}
        \label{subfig:hdr}
    \end{subfigure}
    \begin{subfigure}[b]{0.33\linewidth}
        \includegraphics[width=\linewidth]{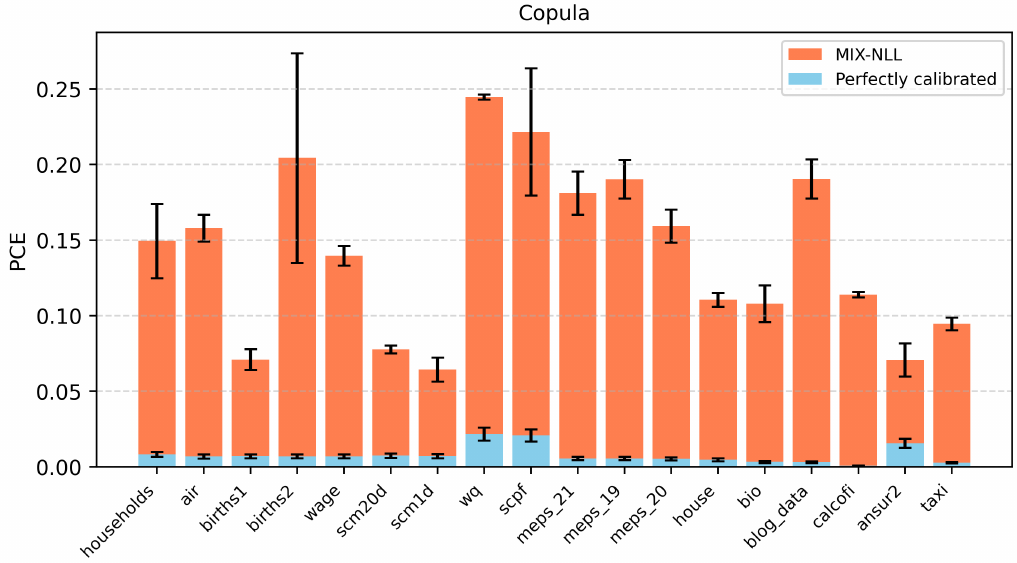}
        \caption{Copula}
        \label{subfig:copula}
    \end{subfigure}
    \caption{PCE values with respect to (a) Location (b) HDR and (c) Copula pre-ranks averaged over five runs across 18 benchmark datasets using the MIX-NLL baseline. Blue bars indicate reference PCE values from a simulated perfectly calibrated model.}
    \label{fig:hyp_test}
\end{figure*}

\section{An Experimental Study of Multivariate Calibration}
\label{sec:motivation}

We conduct a large-scale experimental study to assess the multivariate calibration of (unregularized) multi-output regression models using a set of pre-rank functions introduced in the previous section.

\paragraph{Benchmark Datasets.} Our experiments are performed on 18 real-world multi-output regression datasets drawn from prior work \citep{feldman2022calibratedmultipleoutputquantileregression, wang2022probabilisticconformalpredictionusing, CAMEHL2025105807}. These datasets are widely used in the literature on multivariate calibration \citep{chung2024samplingbased}, conformal prediction \citep{Dheur2025-ou, Guan2021LocalizedCP}, and uncertainty quantification \citep{Angelopoulos2020UncertaintySF}, and serve as a standard benchmark for evaluating calibration methods. We include only datasets with at least 400 training instances and follow the same preprocessing and train-validation-test splitting procedure as in \citet{Dheur2025-ou}. The selected datasets vary in size, containing between 424 and 406,440 training examples. The number of input features $L$ ranges from 1 to 279, and the number of output variables $D$ ranges from 2 to 16.

% Each dataset is randomly split into four parts: [TO FILL]\% training,  [TO FILL]\% validation and [TO FILL]\% test.

% We consider [TO FILL] types of probabilistic neural models:
%\noindent \textbf{Neural Probabilistic Regression Baselines.} As our base probabilistic predictor $F_{\theta}$ we consider a Mixture of $K$ multivariate Gaussian distributions. For each input $x \in \mathcal{X}$ and the mixture component $k \in [K]$ the network predicts the mixture weights $\pi_k(x)$, the mean vector $\mu_k(x) \in \mathbb{R}^D$ and the lower triangular Cholesky factor $L_k(x)$ which is then multiplied by its own transpose to get the covariance matrix $\Sigma_{k}(x) \in \mathbb{R}^{D \times D}$. This way, the positive semi-definiteness of covariance matrix is ensured. Thus, the predicted conditional density takes the form $ \hat{f}_{Y|X=x} = \sum_{k=1}^K \pi_k(x) \hspace{0.3em} \mathcal{N}(\cdot| \mu_k(x), \Sigma_k(x))$ where $\pi_k(x) \geq 0$ and $\sum_{k}\pi_k(x) = 1$. We train this model using NLL scoring rule, and will refer to this baseline as MIX-NLL. Further details about the model architecture can be found in the experimental section.

\paragraph{Neural probabilistic regression model.} Our base probabilistic predictor models a conditional predictive distribution as a mixture of $K$ multivariate Gaussian components, where all parameters are generated by a hypernetwork. For each input $x \in \mathcal{X}$ and each mixture component $k \in [K]$, the network predicts the mixture weight $\pi_k(x)$, the mean vector $\mu_k(x) \in \mathbb{R}^D$, and the lower triangular Cholesky factor $L_k(x)$. The covariance matrix is then computed as $\Sigma_k(x) = L_k(x) L_k(x)^\top$, ensuring positive semi-definiteness by construction. The resulting conditional density takes the form: $\hat{f}_{Y|X=x} = \sum_{k=1}^K \pi_k(x) \, \mathcal{N}(\cdot \mid \mu_k(x), \Sigma_k(x))$ where $\pi_k(x) \geq 0$ and $\sum_{k=1}^K \pi_k(x) = 1$. We train this model using the NLL scoring rule and refer to this baseline as MIX-NLL. Further architectural and training details are provided in the Experiments section.

% \subsection{Experimental Setup}

% The number of examples ranges from [TO FILL] to [TO FILL], and the input dimensionality varies between [TO FILL] and [TO FILL]. All neural models are trained using a fully connected architecture with [TO FILL] layers and [TO FILL] hidden units per layer, [TO FILL] activations, and the Adam optimizer.

% For each trained model, we evaluate multivariate calibration on the test set using several pre-rank projection functions (Section~\ref{sec:pre-ranks}): marginal projections (1D), linear projections (random directions), PCA projections, and HDR-based pre-ranks. For each pre-rank $\rho$, we compute the projected target variable $T = \rho(X,Y)$ and estimate the projected conditional CDF $\hat{F}_{T|X}$ using $N' = [TO FILL]$ Monte Carlo samples from the model. We then calculate the projected PIT value $\hat{F}_{T|X}(T)$ and use it to assess univariate calibration via the PIT histogram and a smoothed calibration error.

\paragraph{Results.}
Figure~\ref{fig:hyp_test} reports the test PCE values for the location, HDR, and copula pre-rank functions, averaged over five independent runs (corresponding to different train-validation-test splits) on each of the 18 benchmark datasets using the MIX-NLL model. For reference, we also simulate ideal PCE scores by sampling from a perfectly calibrated model; these reference values are shown in blue. Due to space constraints, we display results for only a subset of pre-rank functions; figures for the remaining ones are provided in the Appendix. As shown, the MIX-NLL model exhibits substantial miscalibration across all pre-ranks and the majority of datasets. 

We assess the significance of PCE values by generating $5 \times 10^4$ samples of uniformly distributed PITs to approximate the null distribution under perfect calibration for each dataset and pre-rank. One-sided p-values (Holm-corrected) show that all deviations are statistically significant, confirming systematic miscalibration (see Appendix).

In summary, these results highlight that despite being trained with a strictly proper scoring rule, MIX-NLL exhibits significant miscalibration across multiple pre-rank functions on standard benchmarks. In the following section, we investigate how calibration can be improved for these pre-ranks.

% A Pre-rank Regularization method for Multivariate Calibration

% \section{A Pre-rank-guided Regularization Method }\label{sec:methodology}

\section{A Pre-rank Regularization Framework }\label{sec:methodology}
%While proper scoring rules implicitly reward calibrated predictions when evaluating predictive performance, minimizing such a rule during training does not, in general, guarantee calibrated outputs. Indeed, optimizing a strictly proper scoring rule under model misspecification may yield predictions that are sharp but miscalibrated. This stems from the fact that proper scoring rules combine two objectives, calibration and sharpness, and do not penalize miscalibration explicitly. Decompositions of scoring rules \citep{Brocker2008-bh} show that accuracy can increase at the expense of calibration.

%In this section, we describe how calibration can be enforced during training by augmenting the loss function with a dedicated penalty term. Our approach builds upon pre-rank functions introduced earlier to define a class of \emph{projected PITs}, which reduce multivariate calibration to the univariate case. We first explain how these quantities are used to assess calibration, then describe how they are incorporated into a regularized training objective.

Although proper scoring rules are designed to reward calibration, minimizing them during training does not guarantee that the resulting models will be calibrated. Under model misspecification, even strictly proper scoring rules may favor sharp yet miscalibrated predictions, as they do not explicitly penalize miscalibration \citep{Brocker2008-bh}.

In this section, we introduce a training strategy that explicitly enforces calibration by augmenting the loss function with a calibration-specific regularization term. Building on the pre-rank functions introduced earlier, we leverage \emph{projected PITs} to reduce the multivariate calibration problem to a collection of univariate calibration tasks.

\noindent \textbf{Calibration of Projected PITs.}
As stated in Definition~\ref{Def1}, assessing calibration requires access to the conditional CDF $\hat{F}_{T|X}$. Since this CDF is typically unavailable in closed form, we approximate it empirically. For a given test point $(X_i, Y_i)$, we draw $S$ samples $\hat{Y}_{1}, \dots, \hat{Y}_{S} \sim \hat{F}_{Y|X = X_i}$, and compute the projected values:
$$
T_{i} = \rho(X_i, Y_i) \quad \text{and} \quad \hat{T}_{s} = \rho(X_i, \hat{Y}_{s}) \quad \text{for } s = 1, \dots, S.
$$
The conditional CDF $\hat{F}_{T|X=X_i}$ is then estimated using a smoothed indicator function:
\begin{equation}
    \hat{F}_{T|X=X_i}(t) = \frac{1}{S} \sum_{s=1}^{S} \mathbf{1}_\tau(\hat{T}_s \leq t),
    \label{eq:projected-pits}
\end{equation}
where $\mathbf{1}_\tau(x \leq y) = \sigma(\tau(y - x))$, and $\sigma(z) = \frac{1}{1 + e^{-z}}$ is the sigmoid function with temperature parameter $\tau$. The \emph{projected PIT} is defined as the value of this estimated CDF evaluated at the true projected target:
\begin{equation}
Z = \hat{F}_{T|X}(T). \label{eq:projPIT}
\end{equation}
Under perfect calibration for the pre-rank function $\rho$, the projected PIT values $Z$ should follow a uniform distribution in $[0,1]$.

As in \eqref{eq:pce}, under a given pre-rank function \( \rho \), the PCE can be used to quantify the deviation of projected PIT values from uniformity. This extends the univariate PCE formulation to the multivariate setting by applying it to any scalar projection of the multivariate predictions, thereby enabling the assessment of calibration with respect to a chosen pre-rank. Note, however, that the empirical CDF of the projected PITs is not differentiable and therefore cannot be used directly in gradient-based training. To address this, we rely on differentiable approximations that enable calibration to be enforced during model training.

% Enforcing Multivariate Calibration via Regularization

\paragraph{Pre-Rank calibration via Regularization.} Recall from \eqref{eq:projPIT} that the projected PIT variable $Z$ depends on a chosen pre-rank function $\rho$. To encourage calibration with respect to $\rho$, we define a differentiable regularizer based on the PCE-KDE expression in \eqref{PCE-KDE-uni}, using the projected PIT values.

Following \citet{Wilks2018-jz} and \citet{Dheur2024-lw}, we augment the training loss with a differentiable penalty that steers the model toward improved calibration during training. Specifically, the augmented objective is:
\begin{equation}
\mathcal{L}(\theta; \mathcal{D}) = \frac{1}{N} \sum_{i=1}^N S(F_\theta(x_i), y_i) + \lambda \, \mathcal{R}_{\text{PCE-KDE}}(\theta; \mathcal{D}; \rho),
\label{eq:train-objective}
\end{equation}
where $S$ is a strictly proper scoring rule and $\lambda \geq 0$ controls the strength of the calibration regularization. The case $\lambda = 0$ recovers standard unregularized training, while increasing $\lambda$ prioritizes calibration, potentially at the expense of predictive accuracy. Note that the regularizer $\mathcal{R}_{\text{PCE-KDE}}(\theta; \mathcal{D}; \rho)$ is specific to the chosen pre-rank function $\rho$. We refer to the resulting method as \textbf{pre-rank}.

\begin{figure*}[!htbp]
    \centering
    \begin{subfigure}[b]{0.32\linewidth}
        \includegraphics[width=\linewidth]{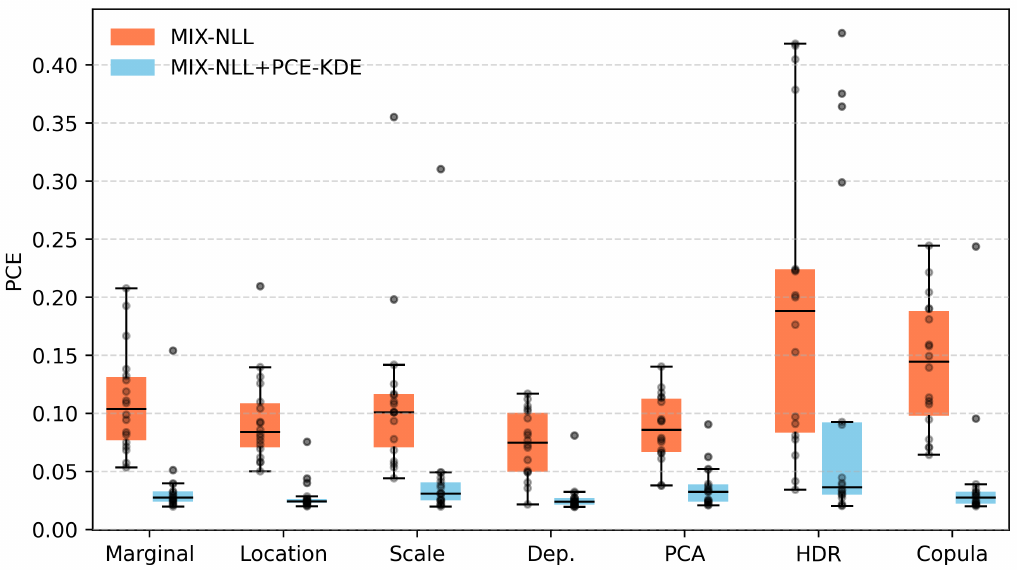}
        \caption{PCE}
        \label{subfig:pce}
    \end{subfigure}
    \begin{subfigure}[b]{0.32\linewidth}
        \includegraphics[width=\linewidth]{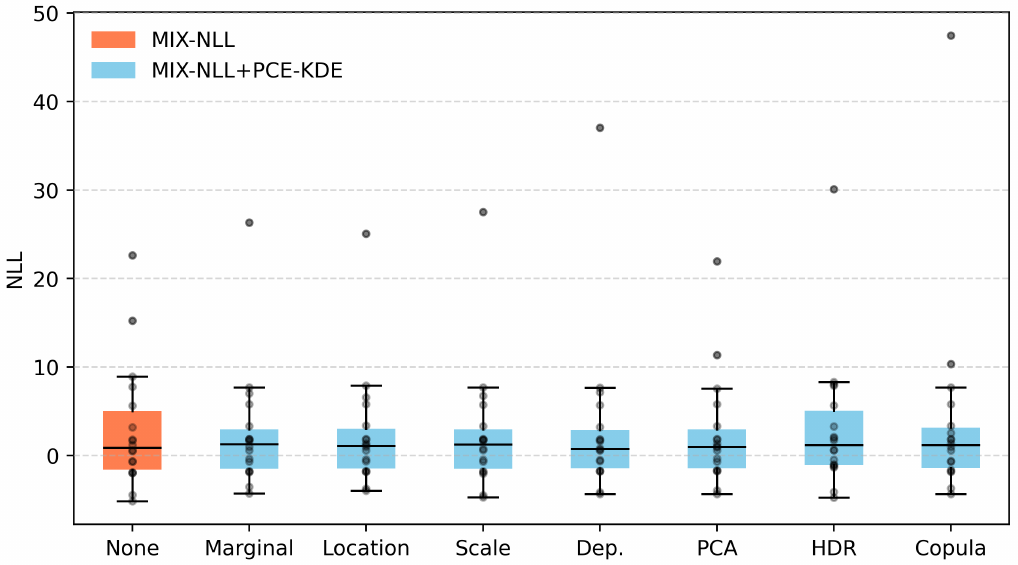}
        \caption{NLL}
        \label{subfig:nll}
    \end{subfigure}
    \begin{subfigure}[b]{0.32\linewidth}
        \includegraphics[width=\linewidth]{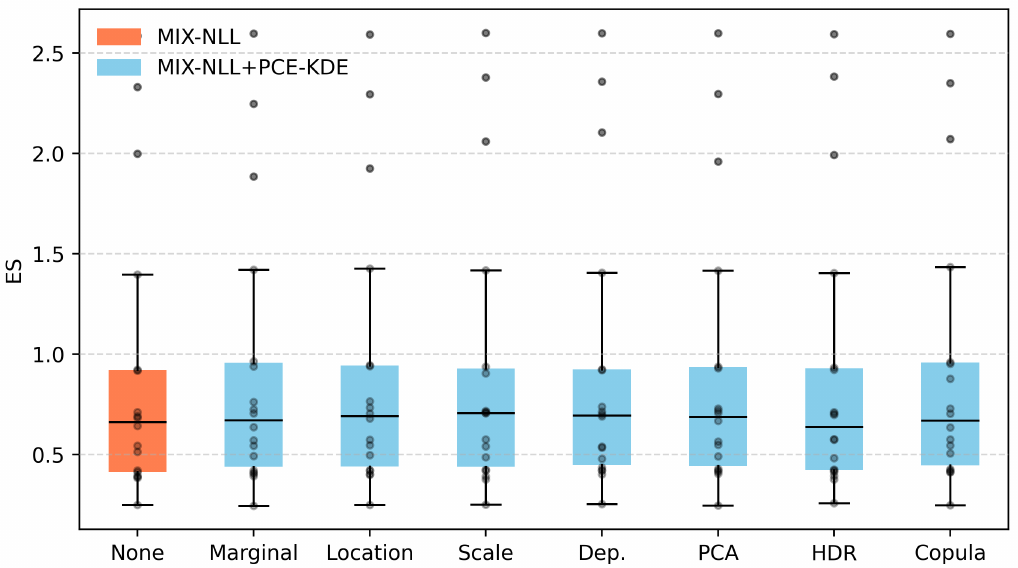}
        \caption{Energy Score}
        \label{subfig:es}
    \end{subfigure}
    \caption{Performance on 18 real multivariate benchmark datasets. Orange: MIX-NLL (no regularization). Blue: MIX-NLL+PCE-KDE (proposed). Metrics are calculated across seven pre-rank functions, and averaged over five runs. In subplots (b) and (c), the “None” box refers to the unregularized MIX-NLL trained without pre-rank.}
    \label{fig:metrics-before-after}
\end{figure*}

\paragraph{Marginal and Pre-rank Calibration.} Calibration with respect to an arbitrary pre-rank does not necessarily imply marginal calibration for each output dimension. However, ensuring marginal calibration is crucial, as any multivariate distribution can be decomposed into its marginal distributions and a dependence structure, according to Sklar's theorem \citep{sklar1959fonctions}. An important exception is the copula pre-rank, which is designed to capture both marginal and joint miscalibration. Since this property does not hold for many commonly used pre-ranks, we propose to explicitly account for marginal calibration alongside any chosen pre-rank.
% Ensuring both marginal and pre-rank calibration is essential, as a model that appears calibrated under a pre-rank may still be miscalibrated if its marginals are inaccurate. 
To this end, we define a combined regularizer:
\begin{equation}
    \frac{1}{D} \sum_{d=1}^D \mathcal{R}_{\text{PCE-KDE}}(\theta; \mathcal{D}; \rho_{\text{marg}}^d) + \mathcal{R}_{\text{PCE-KDE}}(\theta; \mathcal{D}; \rho),
\end{equation}
where $\rho_{\text{marg}}^d$ denotes the marginal pre-rank function for the $d$-th output dimension.

As before, any training loss can be augmented with this combined regularizer. We refer to the resulting model variant as \textbf{marginal+pre-rank}. This formulation is designed to enforce marginal calibration without compromising, and potentially enhancing, calibration along the selected pre-rank direction.

% To do that, PCA pre-rank projects the outcome onto directions of highest variance. These directions are given by the top principal components of the predicted distribution, yielding the following pre-rank function:
%The result is a collection of orthonormal directions \( (V_1(x), \dots, V_D(x)) \) aligned with the major axes of variation in the model’s predictions.
%This construction enables calibration assessment along directions that are statistically meaningful, as they reflect the principal modes of uncertainty in the predictive model.

\paragraph{A PCA-Based Pre-rank and Regularizer.} Given the multivariate nature of the output, we propose a novel pre-rank function based on principal component analysis (PCA). The goal is to assess calibration along the directions of highest variance in the predictive distribution. Specifically, the PCA pre-rank projects the output onto the top principal components of the model's predictive covariance, yielding the following function:
$$
\rho_{\text{pca}}^d(x, y) = y \cdot V_d(x),
$$
where $V_d(x) \in \mathbb{R}^D$, for $d \in \{1, \ldots, D\}$, denotes the $d$-th principal component of the covariance matrix associated with the predicted conditional distribution $\hat{F}_{Y|X=x}$. These components are obtained by sampling from the model’s predictive distribution and performing PCA on the resulting samples. 
% Because the components depend solely on the model's predicted distribution $\hat{F}_{Y|X}$, and not on the true observations, they offer a consistent, model-specific basis for evaluating and enforcing calibration.

While our PCA pre-rank can be treated like any other pre-rank function, it offers the additional advantage of dimensionality reduction. Its computational complexity is \(O(SD^2 + D^3)\), but when only the top principal components \(d^*\) are retained, the combined PCA + pre-rank cost scales with \(d^*\) instead of \(D\). This is particularly beneficial in high-dimensional settings, where evaluating calibration across all marginals can be computationally intensive and statistically unstable.
%To address this, we leverage PCA not only as a pre-rank but also as a dimensionality reduction tool to support pre-rank-based calibration. 

To this end, we project the outputs onto the top $d^*$ principal components that explain a large proportion of the predictive variance (e.g., 80\%). We then compute PCEs along these components and combine them with the regularization term from an arbitrary pre-rank $\rho$, yielding the following combined regularizer:
$$
\frac{1}{d^*} \sum_{d=1}^{d^*} \mathcal{R}_{\text{PCE-KDE}}(\theta; \mathcal{D}; \rho_{\text{pca}}^d) + \mathcal{R}_{\text{PCE-KDE}}(\theta; \mathcal{D}; \rho).
$$
We refer to this approach as \textbf{PCA+pre-rank}. 

To clarify when different pre-rank calibration notions are interchangeable, we provide a sufficient condition under which calibration with respect to one pre-rank function implies calibration with respect to another.

% \begin{figure*}[t!]
%     \centering
%     \includegraphics[width=\textwidth]{Figures/AAAI-only-figures/pce_comparison_grid.pdf}
%     \caption{Mean PCE and standard error over 5 runs for different datasets. Each subplot shows the PCE computed with a specific pre-rank function used to evaluate multivariate calibration.}
%     \label{fig:hyp_test}
% \end{figure*}
\paragraph{Equivalence of Pre-Rank Calibration.}
% To clarify when different pre-rank calibration criteria can be considered interchangeable, we provide a sufficient condition under which calibration with respect to one projection function implies calibration with respect to another.
Let $\rho_1$ and $\rho_2$ be two projection functions. We say that the calibration criteria associated with $\rho_1$ and $\rho_2$ are \textit{equivalent} if a model is calibrated with respect to $\rho_1$ if and only if it is calibrated with respect to $\rho_2$. The following proposition characterizes a sufficient condition for such equivalence:

\begin{proposition}
\label{prop:monotonic_equivalence}
\emph{For every fixed $x \in \mathbb{R}^L$, the function $y \mapsto \rho_2(x, y)$ must be a strictly monotonic bijective transformation of $y \mapsto \rho_1(x, y)$. That is, there exists a strictly increasing or decreasing bijection $h_x$ such that for all $y \in \mathbb{R}^D$,}
$$
\rho_2(x, y) = h_x(\rho_1(x, y)).
$$
\end{proposition}
The full proof is provided in the Appendix. This result shows that strictly monotonic transformations of projection functions preserve the distribution of PIT values, and therefore, the notion of calibration. However, such conditions are rarely met in practice, and different pre-rank functions often lead to distinct, potentially incompatible calibration assessments.

\section{Experiments}\label{sec:experiments}

We extend our earlier empirical analysis to evaluate the effectiveness of the proposed pre-rank regularization framework. Specifically, we train the MIX-NLL model with a PCE-KDE regularizer applied to the following pre-rank functions: (1) marginal, (2) location, (3) scale, (4) dependency, (5) PCA, (6) HDR, and (7) copula. We refer to this model as \textbf{MIX-NLL+PCE-KDE} trained on \textbf{pre-rank}. We then compute the PCE values on the test set and compare them to those obtained from the unregularized MIX-NLL baseline. All experiments are conducted using the same 18 benchmark datasets introduced earlier.

\paragraph{Metrics.} We evaluate model performance using three metrics: PCE (as a measure of calibration), negative log-likelihood (NLL), and energy score (ES). Details on the empirical computation of ES are provided in the Appendix.

\paragraph{Hyperparameters.} For the MIX-NLL baseline, we use a mixture of $K = 5$ multivariate Gaussian components. The neural network consists of three fully connected layers with 100 hidden units each, ReLU activations, and is trained using the Adam optimizer with a learning rate of $10^{-4}$. To compute the PCE-KDE regularizer, we estimate the projected PITs $\hat{F}_{T|X}(T)$ using $S = 100$ samples drawn from the predictive distribution with the parameters set to $p=1$ and $M=100$. The temperature parameter $\tau$ in the smoothed indicator function is set to $100$, following prior work in \citet{Dheur2023-it}. The regularization strength $\lambda$ in~\eqref{eq:train-objective} controls the degree of calibration enforcement with respect to the chosen pre-rank. As observed in prior work \citep{karandikar2021softcalibrationobjectivesneural, Wessel2025-je}, increasing $\lambda$ typically improves calibration (lower PCE) but may degrade predictive performance (higher NLL or ES). Following the tuning strategy used in \citet{karandikar2021softcalibrationobjectivesneural} and \citet{Dheur2023-it}, we select $\lambda$ to minimize PCE while ensuring that ES does not increase by more than 10\% relative to the best ES obtained when $\lambda = 0$. This strategy allows us to improve calibration without sacrificing predictive accuracy. The optimal $\lambda$ is tuned on validation set and selected from $\{0, 0.01, 0.1, 1, 5, 10\}$ for each (dataset, pre-rank) pair. The exact values of selected $\lambda$ are reported in the Appendix.

% \begin{figure}[!htbp]
% \centering
% \includegraphics[width=0.47\textwidth]{Figures/AAAI-only-figures/box_plot_after_pce.png} % Reduce the figure size so that it is slightly narrower than the column.
% \caption{PCE values across 21 real-world datasets, computed using seven prerank functions after applying PCE-KDE regularization with MIX-NLL model.}
% \label{fig:pce-after-reg}
% \end{figure}

% \noindent \textbf{Comparison of NLL and ES} 

\section{Results}

Figure~\ref{subfig:pce} compares the test PCE of the unregularized MIX-NLL model with the regularized version, MIX-NLL+PCE-KDE, where calibration is explicitly enforced with respect to each pre-rank function. As expected, regularization substantially reduces the PCE for the corresponding pre-rank. For all pre-ranks, the median PCE across datasets is consistently lower after regularization. Additionally, the distribution of PCE values (illustrated via box plots) becomes noticeably tighter, indicating that the regularization leads to more consistent calibration improvements across datasets. Full results, averaged over five runs for each dataset and pre-rank, are provided in the Appendix.

Despite the overall improvements, calibration remains challenging for a few datasets with initially high PCE values, particularly under the HDR pre-rank. As shown in Figure~\ref{subfig:pce}, four datasets exhibit little to no improvement in PCE when regularized with HDR. This aligns with the limitation highlighted by \citet{chung2024samplingbased}: when the model’s predictive distribution poorly approximates the true one, HDR recalibration struggles to recover the underlying dependency structure among target variables. Consequently, the effectiveness of HDR as a pre-rank is highly sensitive to model specification. In these cases, the underlying Mixture of Gaussians model may be misaligned with the data distribution.

Figures~\ref{subfig:nll} and~\ref{subfig:es} show the NLL and ES values after applying pre-rank-based regularization, alongside their values without regularization (denoted ``None''). As the results indicate, regularization does not significantly degrade predictive performance. This demonstrates that enforcing calibration through pre-rank regularization maintains predictive quality. Importantly, the regularization strength $\lambda$ is selected to control increases in ES, ensuring that improvements in calibration do not come at the expense of accuracy.

\begin{figure}[!t]
    \centering
    \begin{subfigure}[b]{0.32\linewidth}
        \includegraphics[width=\linewidth]{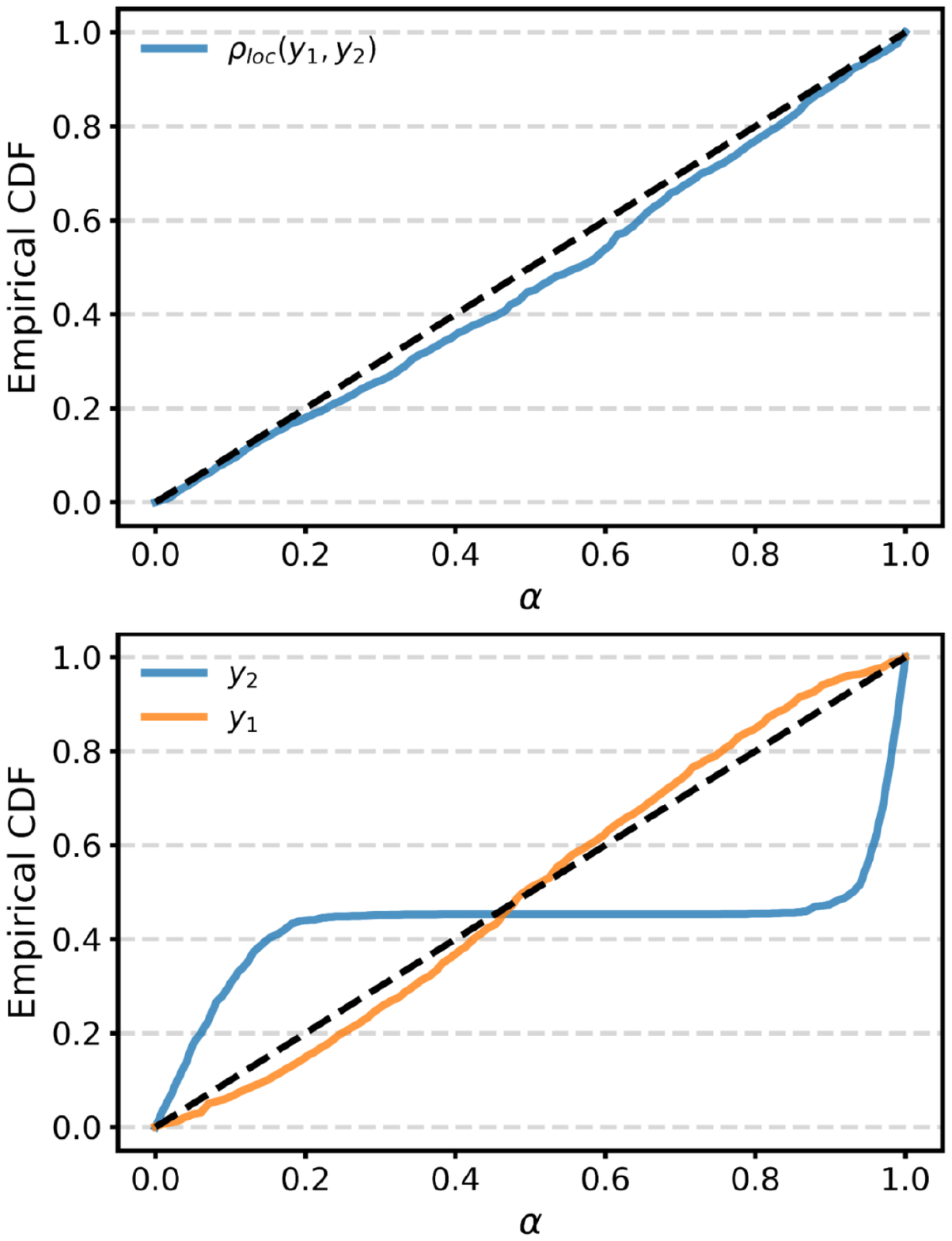}
        \caption{Location}
        \label{subfig:loc-prerank}
    \end{subfigure}
    \begin{subfigure}[b]{0.32\linewidth}
        \includegraphics[width=\linewidth]{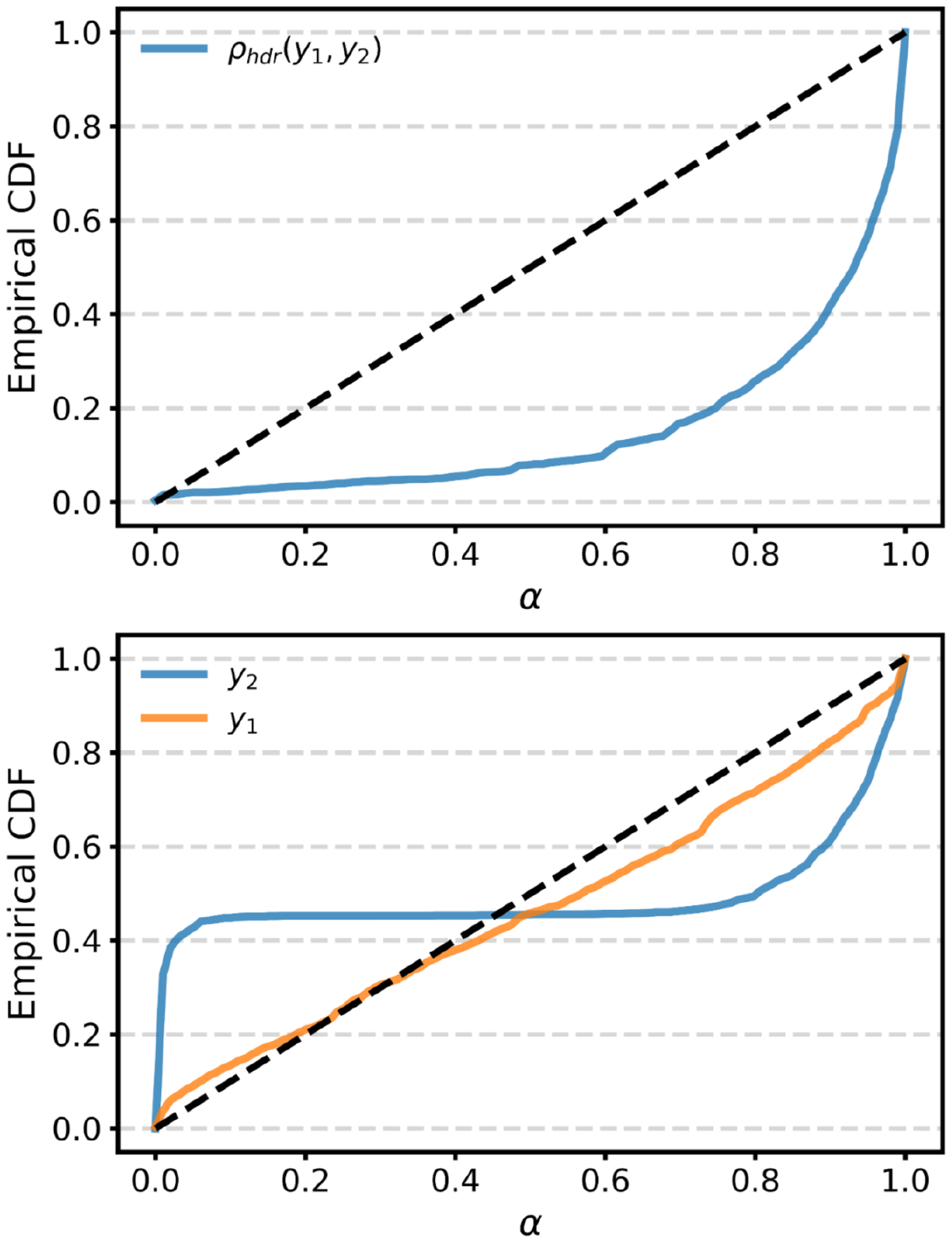}
        \caption{HDR}
        \label{subfig:hdr-prerank}
    \end{subfigure}
    \begin{subfigure}[b]{0.32\linewidth}
        \includegraphics[width=\linewidth]{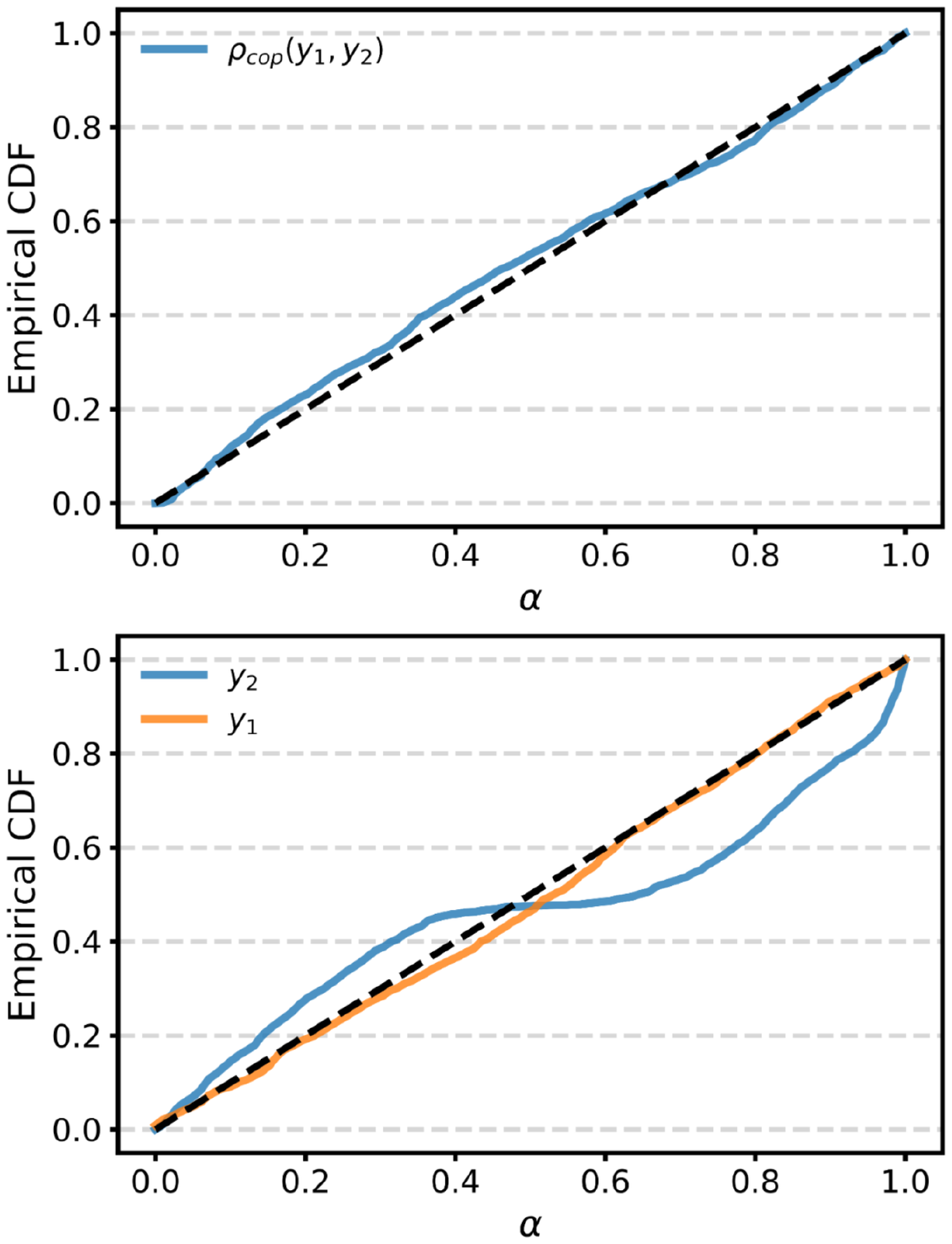}
        \caption{Copula}
        \label{subfig:cop-prerank}
    \end{subfigure}
    \caption{Reliability plots on \texttt{wage} dataset using MIX-NLL + PCE-KDE on prerank. Top row: calibration curves with respect to (a) Location, (b) HDR, and (c) Copula preranks. Bottom row: corresponding marginal calibration curves.}
    \label{fig:rel-plots-prerank}
\end{figure}

\begin{figure}[!t]
    \centering
    \begin{subfigure}[b]{0.32\linewidth}
        \includegraphics[width=\linewidth]{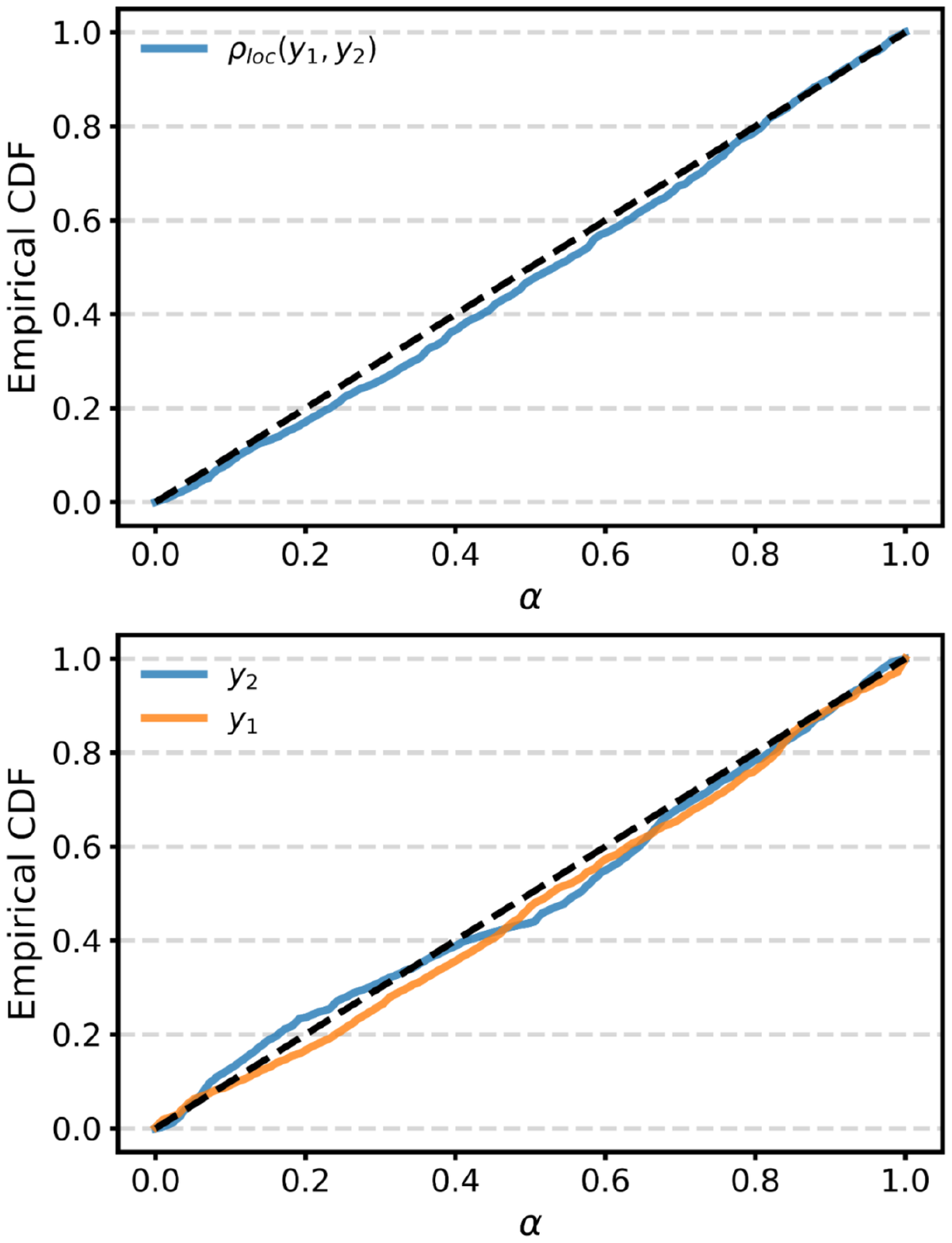}
        \caption{Location}
        \label{subfig:loc-marg-prerank}
    \end{subfigure}
    \begin{subfigure}[b]{0.32\linewidth}
        \includegraphics[width=\linewidth]{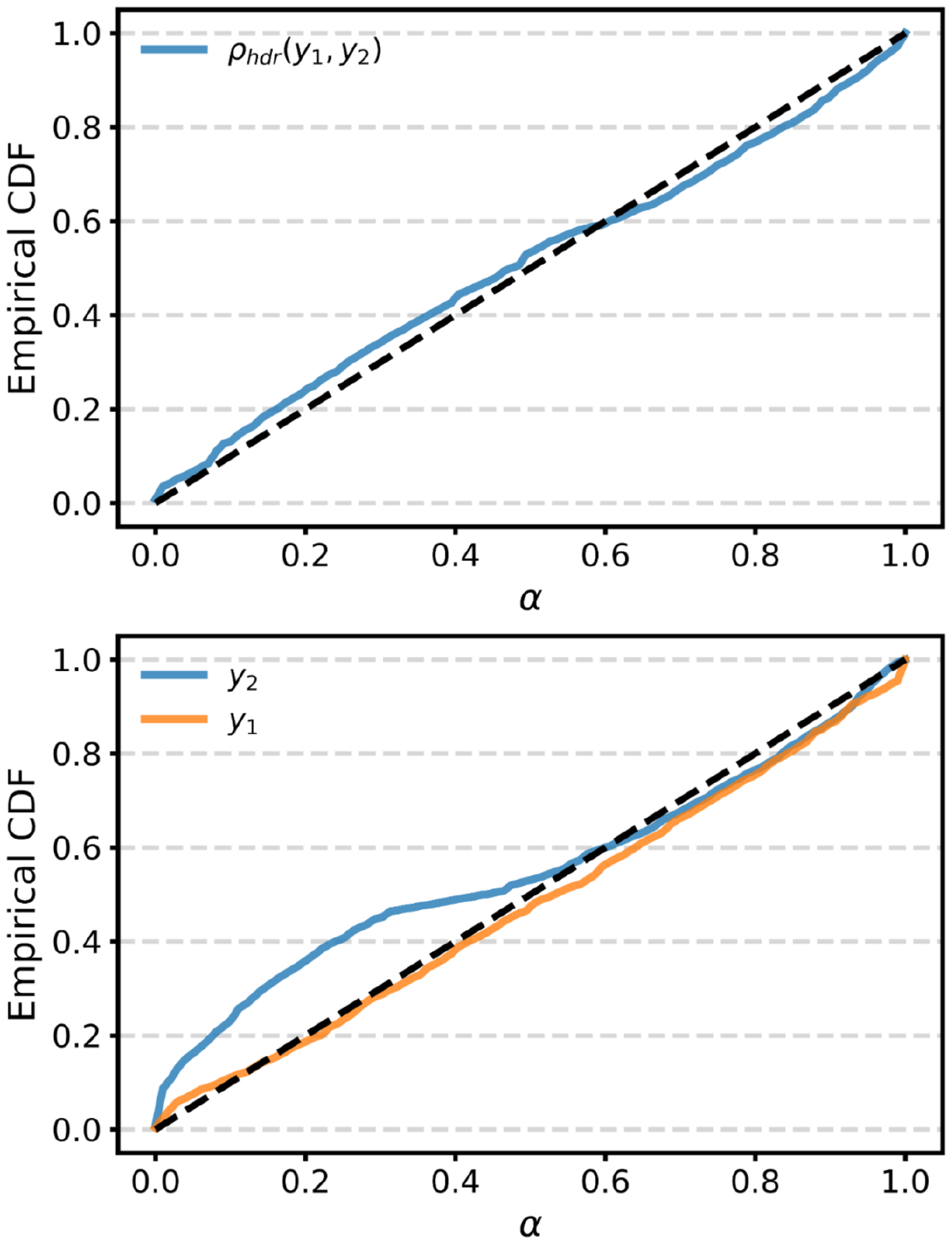}
        \caption{HDR}
        \label{subfig:hdr-marg-prerank}
    \end{subfigure}
    \begin{subfigure}[b]{0.32\linewidth}
        \includegraphics[width=\linewidth]{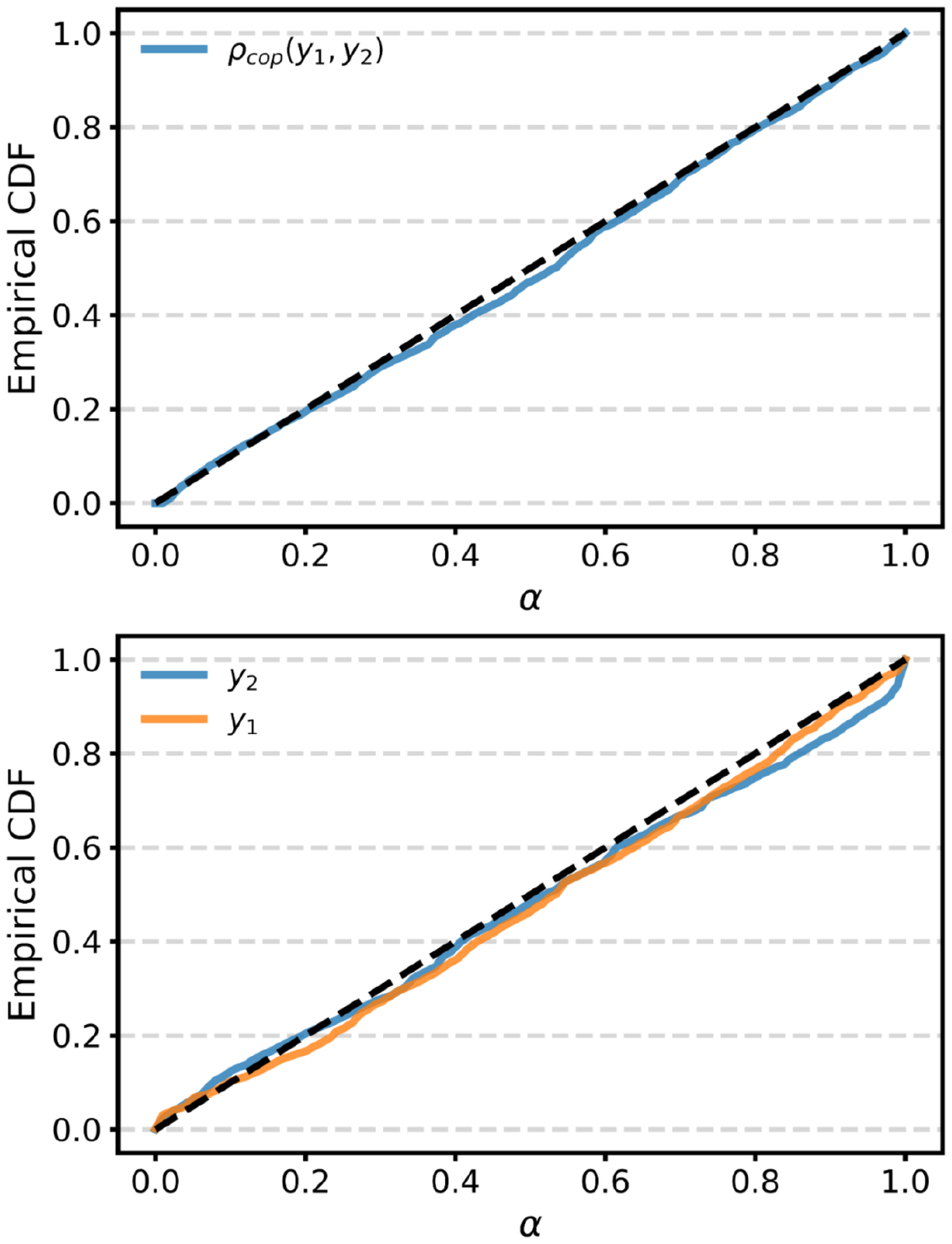}
        \caption{Copula}
        \label{subfig:cop-marg-prerank}
    \end{subfigure}
    \caption{Reliability plots on \texttt{wage} dataset using MIX-NLL + PCE-KDE on marginal+pre-rank. Top row: calibration curves with respect to (a) Location, (b) HDR, and (c) Copula pre-ranks. Bottom row: corresponding marginal calibration curves.}
    \label{fig:rel-plots-marg-prerank}
\end{figure}

\paragraph{Marginal and Pre-rank Calibration.} The reliability plots in Figure~\ref{fig:rel-plots-prerank} show that regularizing solely with respect to a specific pre-rank improves calibration for that pre-rank, but not necessarily for the marginal distributions. Among the three examples shown, only the copula pre-rank also improves marginal calibration, as it is designed to capture both marginal and joint structure. A similar effect is observed with the PCA pre-rank (see Appendix), which improves marginal calibration by projecting prediction-observation pairs onto all principal components and averaging the resulting PCEs, acting as a marginal pre-rank in a rotated space. In contrast, HDR-only regularization shows minimal improvement on the \texttt{wage} dataset (Figure~\ref{subfig:hdr-prerank}), likely due to its dependence on the quality of the model’s predictive distribution. By comparison, Figure~\ref{subfig:hdr-marg-prerank} demonstrates that combining marginal and pre-rank regularization leads to improvements in both HDR and marginal calibration. This is likely because the marginal PCE provides a complementary signal that is less sensitive to model misspecification, thereby supporting more reliable calibration across both marginal and structured components.

% Figure \ref{fig:rel-plots-prerank} shows that regularizing on a specific prerank improves calibration for that prerank, but not necessarily for marginals. Reliability curves for other preranks are provided in the Appendix. Among the three shown in Figure \ref{fig:rel-plots-prerank}, only the Copula prerank improves marginal calibration, as it is explicitly designed to capture both marginal and joint dependencies.
% \noindent Interestingly, we also observe in the Appendix that the PCA prerank improves marginal calibration. This occurs because it projects prediction–observation pairs along directions of maximal variance, computes PCEs along these axes, and averages them as the regularization objective. This effectively behaves like a marginal prerank in a rotated space, making the improvement in marginal calibration expected. 

% As shown in Figure \ref{fig:rel-plots-marg-prerank}, marginal+pre-rank approach successfully improves calibration with respect to marginals and pre-ranks simultaneously.
% For the HDR prerank, HDR-only regularization yields minimal improvement on the \texttt{wage} dataset (from Figure \ref{subfig:hdr-prerank}), likely due to the limitation discussed earlier. In contrast, marginal+prerank improves both HDR and marginal calibration. We hypothesize this is because HDR is sensitive to the quality of predictive distribution, and including marginal PCE introduces a complementary correction signal that doesn't directly depend on the quality of predictive distribution, guiding the training toward better overall calibration.

% To test whether similar results can be achieved using fewer principal components, instead of the full target dimensionality used by the marginal+prerank method, we conducted additional experiments. 

\paragraph{PCA and Pre-rank Calibration.} Table~\ref{tab:marg-prerank-pca-prerank} reports both marginal and pre-rank PCEs for four training variants: None (no regularization), pre-rank (regularized on a certain pre-rank only), marginal+pre-rank, and PCA+pre-rank. Results are shown for the \texttt{scm1d} dataset with $D = 16$ target dimensions. As expected, the marginal+pre-rank variant achieves the lowest marginal PCEs. However, PCA+pre-rank performs comparably well: even after reducing the dimensionality from 16 to just 3 principal components, its marginal PCEs remain substantially lower than those of the unregularized (None) model. In terms of pre-rank PCEs, marginal+pre-rank again achieves the best results, but the differences are modest. On average, PCA+pre-rank increases the pre-rank PCE by no more than 5\% compared to marginal+pre-rank, while still improving significantly over the None baseline. This indicates that PCA+pre-rank offers a scalable and effective alternative, achieving near-comparable calibration performance using only a few informative projections. Additional results for datasets with $D \geq 4$ are provided in the Appendix.

\begin{table}[!htbp]
\centering
\scriptsize
\begin{tabular}{l|ccccccc}
\hline
Method & Marg & Loc & Scale & Dep & PCA & HDR & Cop \\
\hline
None & \textbf{0.054} & \textbf{0.058} & \textbf{0.108} & \textbf{0.084} & \textbf{0.038} & \textbf{0.078} & \textbf{0.064} \\
\hline
Marg & \textbf{0.025} & 0.028 & 0.074 & 0.063 & 0.038 & 0.08 & 0.040 \\
Loc & \textbf{0.030} & \textbf{0.024} & 0.087 & 0.072 & 0.040 & 0.080 & 0.046 \\
Scale & \textbf{0.047} & 0.059 & \textbf{0.037} & 0.038 & 0.038 & 0.075 & 0.055 \\
Dep & \textbf{0.059} & 0.071 & 0.078 & \textbf{0.020} & 0.043 & 0.081 & 0.064 \\
PCA & \textbf{0.029} & 0.023 & 0.079 & 0.049 & \textbf{0.035} & 0.085 & 0.04 \\
HDR & \textbf{0.059} & 0.069 & 0.113 & 0.089 & 0.042 & \textbf{0.091} & 0.069 \\
Cop & \textbf{0.035} & 0.04 & 0.085 & 0.066 & 0.04 & 0.083 & \textbf{0.035} \\
\hline
Marg+loc & \textbf{0.024} & \textbf{0.022} & 0.074 & 0.05 & 0.036 & 0.073 & 0.038 \\
Marg+scale & \textbf{0.026} & 0.031 & \textbf{0.054} & 0.05 & 0.038 & 0.082 & 0.044 \\
Marg+dep & \textbf{0.025} & 0.030 & 0.071 & \textbf{0.021} & 0.037 & 0.085 & 0.041\\
Marg+HDR & \textbf{0.033} & 0.037 & 0.094 & 0.059 & 0.035 & \textbf{0.087} & 0.042\\
Marg+Cop & \textbf{0.024} & 0.027 & 0.076 & 0.057 & 0.036 & 0.078 & \textbf{0.028} \\
\hline
PCA+loc & \textbf{0.030} & \textbf{0.023} & 0.081 & 0.074 & 0.040 & 0.093 & 0.043 \\
PCA+scale & \textbf{0.032} & 0.032 & \textbf{0.052} & 0.055 & 0.036 & 0.078 & 0.041 \\
PCA+dep & \textbf{0.030} & 0.025 & 0.074 & \textbf{0.022} & 0.036 & 0.081 & 0.039 \\
PCA+HDR & \textbf{0.043} & 0.044 & 0.115 & 0.087 & 0.038 & \textbf{0.095} & 0.06  \\
PCA+Cop & \textbf{0.036} & 0.043 & 0.077 & 0.105 & 0.046 & 0.101 & \textbf{0.042} \\
\hline
% \multirow{10}{*}{scm20d}
% & None & 0.072 & 0.077 & 0.142 & 0.077 & 0.091 & 0.078 \\
% & Marg & 0.033 &- &- &- &- &- \\
% \cline{2-8}
% & Marg+loc & 0.029 & 0.023 & -& -& -&-\\
% & Marg+scale & 0.027 & -& 0.042& -& -&- \\
% & Marg+dep & 0.037 & -& -& 0.026& -& -\\
% & Marg+HDR & 0.041 & -& -& -&0.088 &- \\
% & Marg+Cop & 0.031 & -& -& -& -& 0.028 \\
% \cline{2-8}
% & PCA+loc & 0.034 & 0.022 & -& -& -& \\
% & PCA+scale & 0.034 & -& 0.045 & -& -&- \\
% & PCA+dep & 0.045 & -& -& 0.025& -&- \\
% & PCA+HDR & 0.049 & -& -& -& 0.088 & - \\
% & PCA+Cop & 0.032 & -& -& -& -& 0.029 \\
% \hline
\end{tabular}
\caption{PCE values averaged over five runs from four model variants:
\textbf{None} (no regularization), \textbf{pre-rank} (regularized on certain pre-rank), \textbf{marg+pre-rank}, and \textbf{PCA+pre-rank}. PCE values are shown across different pre-ranks.}
\label{tab:marg-prerank-pca-prerank}
\end{table}
\section{Conclusion}

Multivariate calibration is often assessed using pre-rank functions--projections that reduce prediction-observation pairs to univariate summaries, such as marginal, location, scale, or dependency-based mappings. In a large-scale empirical study on 18 real-world regression datasets, we show that a standard probabilistic predictor, despite being trained with a strictly proper scoring rule, is consistently miscalibrated across all pre-ranks.

To address this, we propose a differentiable regularization framework that enforces calibration during training by penalizing the deviation between quantile levels and the empirical CDF of projected PITs. The method integrates seamlessly with any scoring-rule-based objective and can be extended to jointly enforce marginal and pre-rank calibration.

We also introduce a PCA-based pre-rank that projects predictions onto principal directions of variance, enabling effective calibration in a lower-dimensional space. Despite using only a few components, PCA+pre-rank achieves calibration performance close to marginal+pre-rank.

Empirical results show that our approach consistently improves calibration without compromising predictive accuracy. Overall, this work offers a practical strategy for enforcing multivariate calibration and opens avenues for integrating projection-based regularization into model training.

\clearpage

\bibliography{mainAAAI}

\clearpage
\appendix

\appendix
\section{Appendix}
\subsection{A. Proofs}

\paragraph{Equivalence of Pre-Rank Calibration}
\begin{proposition}
% \label{prop:monotonic_equivalence}
\emph{For every fixed $x \in \mathbb{R}^L$, the function $y \mapsto \rho_2(x, y)$ must be a strictly monotonic bijective transformation of $y \mapsto \rho_1(x, y)$. That is, there exists a strictly increasing or decreasing bijection $h_x$ such that for all $y \in \mathbb{R}^D$,}
$$
\rho_2(x, y) = h_x(\rho_1(x, y)).
$$
\end{proposition}
\textit{Proof.} 
Fix $x \in \mathbb{R}^L$ and define $T_1 = \rho_1(x, Y)$ and $T_2 = \rho_2(x, Y) = h_x(T_1)$, where $h_x$ is a strictly monotonic bijection. Let $\hat{F}_{T_1 \mid X= x}$ and $\hat{F}_{T_2 \mid X= x}$ denote the empirical conditional CDFs of $T_1$ and $T_2$, respectively, estimated using the same sample of predicted values $\{\hat{Y}_i\}_{i=1}^{N'}$ drawn from the predictive distribution $\hat{F}_{Y \mid X= x}$.

As explained in the background section, we estimate these conditional CDFs using the empirical estimator. Since this construction is used solely for evaluation, differentiability of the CDF is not required. Then for any $t \in \mathbb{R}$,
\begin{align*}
\hat{F}_{T_2 \mid X= x}(t) 
&=  \frac{1}{N'} \sum_{i=1}^{N'} \mathbf{1}_\tau(\rho_2(x,\hat{Y}_i) \leq t)\\
&=  \frac{1}{N'} \sum_{i=1}^{N'} \mathbf{1}_\tau(\rho_1(x,\hat{Y}_i) \leq h_x^{-1}(t))\\
&=  \hat{F}_{T_1 \mid X= x}(h_x^{-1}(t)).
\end{align*}

Since \( T_2 = h_x(T_1) \) and 
\[
\hat{F}_{T_2 \mid X= x}(t) = \hat{F}_{T_1 \mid X= x}(h_x^{-1}(t)),
\] we have:
\[
\hat{F}_{T_2 \mid X= x}(T_2) = \hat{F}_{T_1 \mid X= x}(h_x^{-1}(T_2)) = \hat{F}_{T_1 \mid X= x}(T_1),
\] where we used the fact that \( T_1 = h_x^{-1}(T_2) \) by construction. It follows that the PIT value computed under \( \rho_2 \) coincides with the one computed under \( \rho_1 \):
\[
U_2 := \hat{F}_{Z_2 \mid X= x}(Z_2) = \hat{F}_{Z_1 \mid X= x}(Z_1) =: U_1.
\] It follows that \( U_1 \) and \( U_2 \) have the same distribution. In particular,
\[
U_2 \sim \mathcal{U}[0,1] \quad \Longleftrightarrow \quad U_1 \sim \mathcal{U}[0,1],
\]
which establishes the equivalence of the two calibration criteria under the assumed transformation.

\subsection{B. Additional computational details}

\noindent \textbf{Practical note on Copula-based Pre-Rank.}
When using copula-based pre-ranks, one requires access to the joint CDF \( \hat{F}_{Y|X}(y) \), i.e., the probability that all components of \( Y \) are less than or equal to \( y \) given \( X \). However, many models provide only the conditional density \( \hat{f}_{Y|X} \).

We approximate the joint CDF via Monte Carlo sampling. Given input \( X_i \) and target \( Y_i \), we draw \( S \) samples \( \hat{Y}_{i,1}, \dots, \hat{Y}_{i,S} \sim \hat{f}_{Y|X=X_i} \) and estimate:
\begin{equation}
\hat{F}_{Y|X=X_i}(Y_i) \;\approx\; \frac{1}{S} \sum_{s=1}^S \mathbf{1}\left\{ \hat{Y}_{i,s} \leq Y_i \right\},
\end{equation}
where the indicator \( \mathbf{1}\left\{ \hat{Y}_{i,s} \leq y_i \right\} \) is true if and only if \( \hat{Y}_{i,s}^{(d)} \leq y_i^{(d)} \) for all components \( d = 1, \dots, D \).

Since the indicator function is not differentiable, we replace it with a smooth approximate using the sigmoid function \( \sigma(z) = 1/(1 + e^{-z}) \) and a temperature parameter \( \tau > 0 \). This gives:
\begin{equation}
\hat{F}_{Y|X=X_i}(Y_i) \;\approx\; \frac{1}{S} \sum_{s=1}^S \prod_{d=1}^{D} \sigma\left( \tau \Big(Y_i^{(d)} - \hat{Y}_{i,s}^{(d)}\Big) \right),
\end{equation}
where \( y_i^{(d)} \) and \( \hat{Y}_{i,s}^{(d)} \) denote the \( d \)-th components of the vectors \( y_i \) and \( \hat{Y}_{i,s} \), respectively. The product over dimensions enforces that all components of \( \hat{Y}_{i,s} \) fall below the threshold \( y_i \), mimicking the joint indicator condition.

This smooth approximation is fully differentiable with respect to the model parameters (via the samples \( \hat{Y}_{i,s} \)), and thus compatible with gradient-based optimization routines such as backpropagation.

\noindent \textbf{Empirical Calculation of Energy Score}
We use Energy Score (ES) as a scoring rule metric to evaluate our model performance. ES generalizes Continuous Ranked Probability Score (CRPS) to multivariate settings and is computed empirically as:
\begin{equation}
    \text{ES}(\hat{F}, y) = \frac{1}{G} \sum_{i=1}^G \|\hat{Y}_i - y\| - \frac{1}{2G^2} \sum_{i=1}^G \sum_{j=1}^G \|\hat{Y}_i - \hat{Y}_j\|
    \label{eq:energy}
\end{equation}
where $\{\hat{Y}_i\}_{i=1}^G \sim \hat{F}_{Y|X}$ are $G$ samples drawn from the predictive distribution. We set $G=100$ in all experiments.

\subsection{C. Detailed Hypothesis Test Results}
\begin{figure*}[!htbp]
    \centering
    \begin{subfigure}[b]{0.24\linewidth}
        \includegraphics[width=\linewidth]{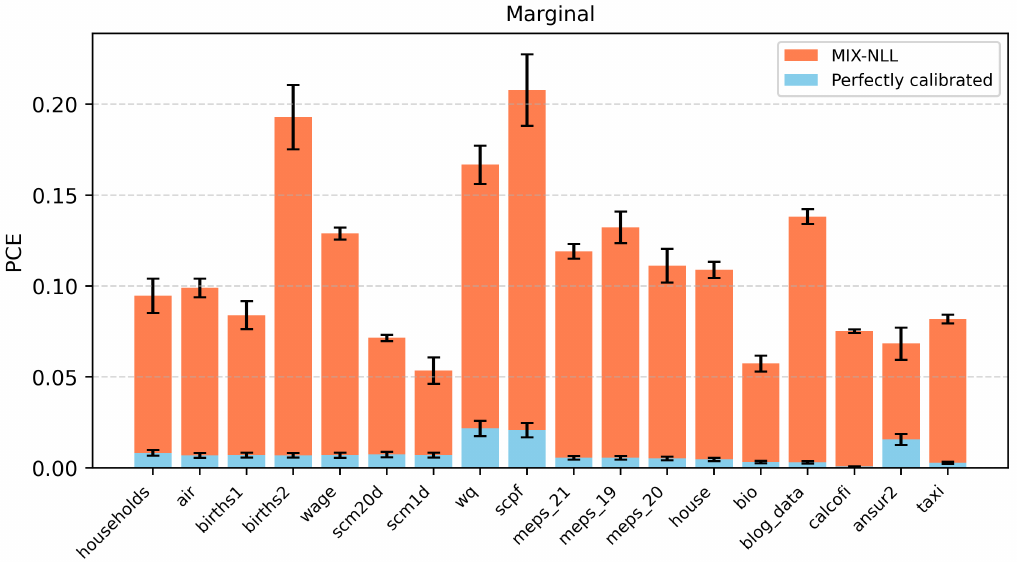}
        \caption{Marginal}
        \label{subfig:marg}
    \end{subfigure}
    \begin{subfigure}[b]{0.24\linewidth}
        \includegraphics[width=\linewidth]{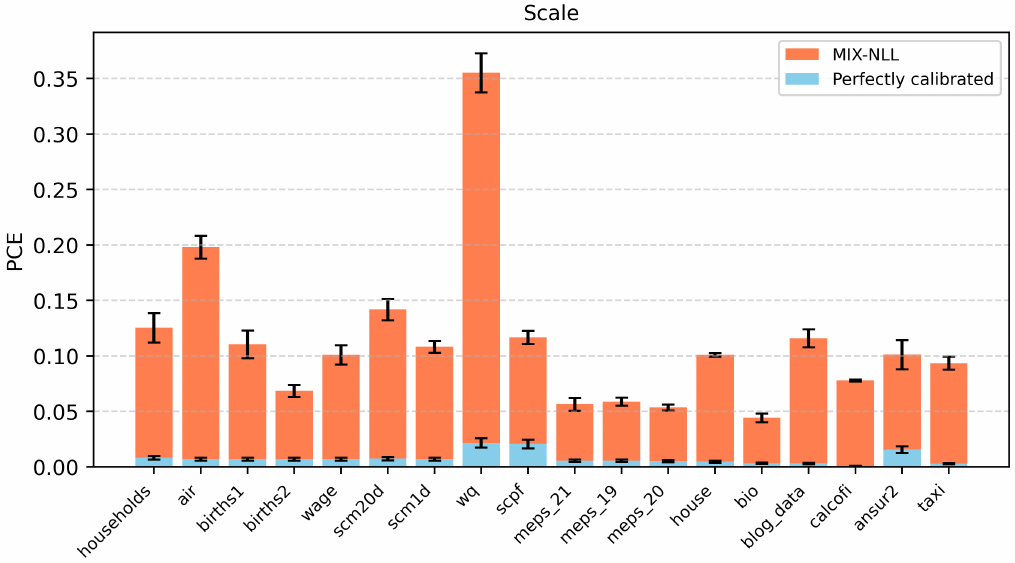}
        \caption{Scale}
        \label{subfig:var}
    \end{subfigure}
    \begin{subfigure}[b]{0.24\linewidth}
        \includegraphics[width=\linewidth]{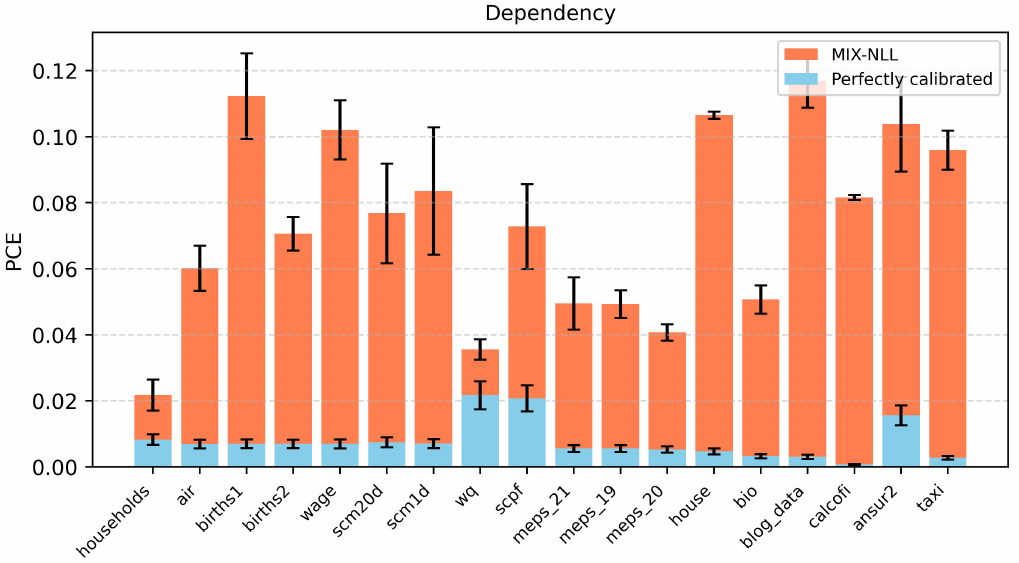}
        \caption{Dependency}
        \label{subfig:dep}
    \end{subfigure}
    \begin{subfigure}[b]{0.24\linewidth}
        \includegraphics[width=\linewidth]{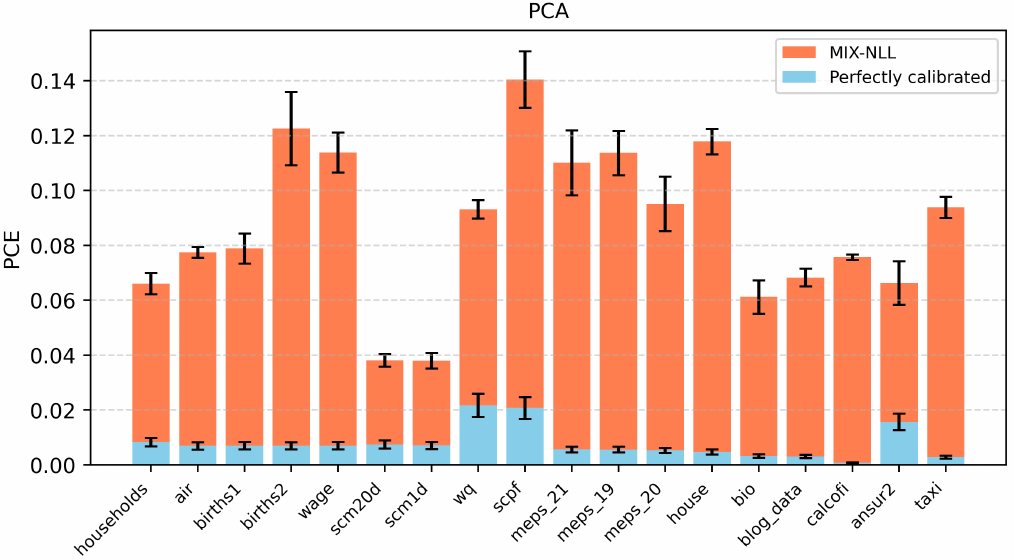}
        \caption{PCA}
        \label{subfig:pca}
    \end{subfigure}
    \caption{PCE values with respect to (a) Marginal (b) Scale (c) Dependency and (d) PCA pre-ranks averaged over five runs across 18 benchmark datasets using the MIX-NLL baseline. Blue bars indicate reference PCE values from a simulated perfectly calibrated model.}
    \label{fig:hyp_test_therest}
\end{figure*}

Figure \ref{fig:hyp_test_therest} shows PCE values for the marginal, scale, dependency, and PCA pre-ranks (excluded from the main paper), averaged over five independent runs on 18 real benchmark datasets using the MIX-NLL model. Simulated PCE scores from a perfectly calibrated model are shown in blue.

\paragraph{Distribution of the Test Statistic.}
To assess the statistical significance of observed PCE values, we estimate the null distribution of the average PCE under perfect calibration for each dataset and pre-rank. For every dataset, \(5 \times 10^4\) samples of the test statistic are generated by simulating independent PIT values uniformly in \([0,1]\), matching the test set size. This captures the variability of the mean PCE expected under ideal calibration.

One-sided $p$-values are computed as the proportion of simulated PCE values exceeding the observed PCE, and Holm correction is applied to control the family-wise error rate across datasets and pre-ranks. After correction, perfect calibration is rejected for all combinations, indicating systematic miscalibration.

\begin{figure*}[t]
    \centering
    \includegraphics[width=\textwidth]{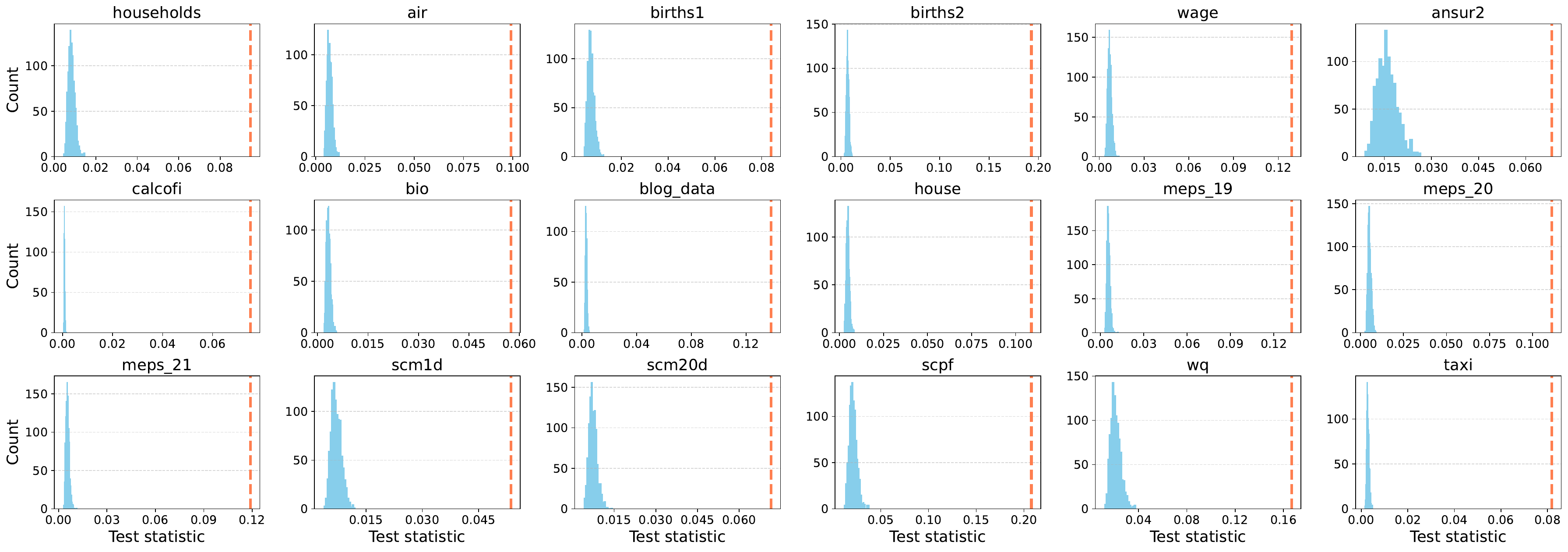}
    \caption{Distributions of the average PCE under the hypothesis of perfect calibration for all datasets, evaluated using the MIX NLL model and the \textbf{marginal} prerank.}
    \label{fig:pce_null_vs_real_marginal}
\end{figure*}

\begin{figure*}[t]
    \centering
    \includegraphics[width=\textwidth]{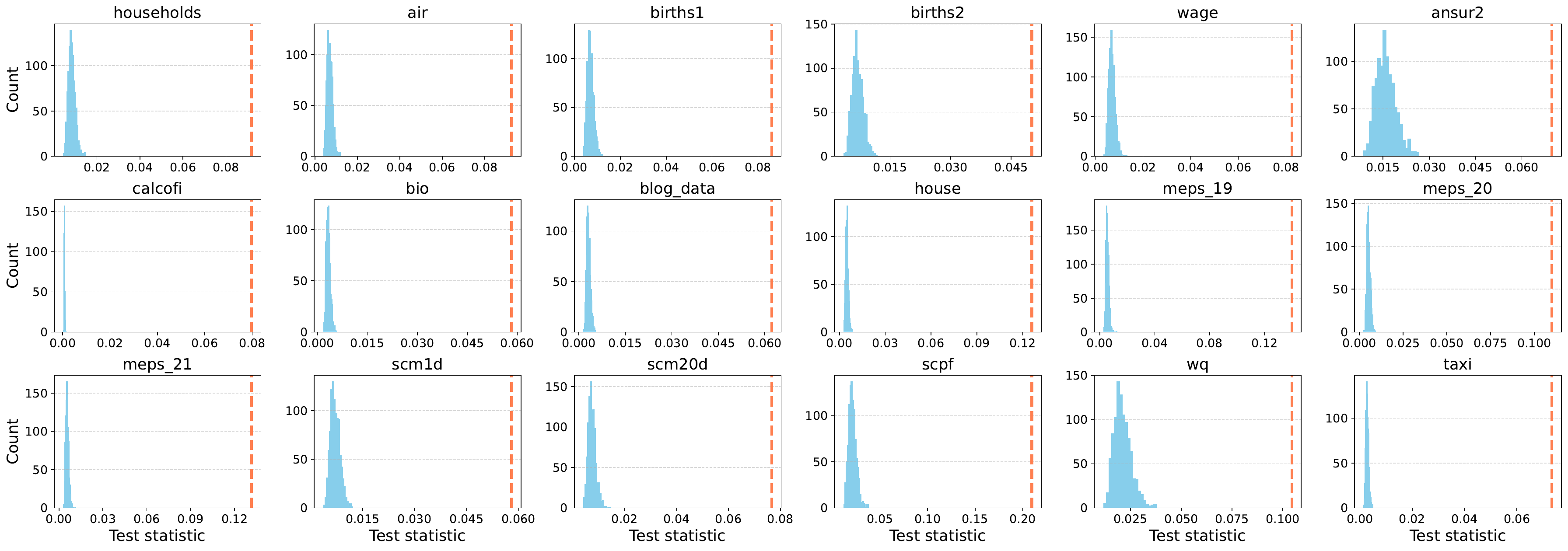}
    \caption{Distributions of the average PCE under the hypothesis of perfect calibration for all datasets, evaluated using the MIX NLL model and the \textbf{location} prerank.}
    \label{fig:pce_null_vs_real_mean}
\end{figure*}

\begin{figure*}[t]
    \centering
    \includegraphics[width=\textwidth]{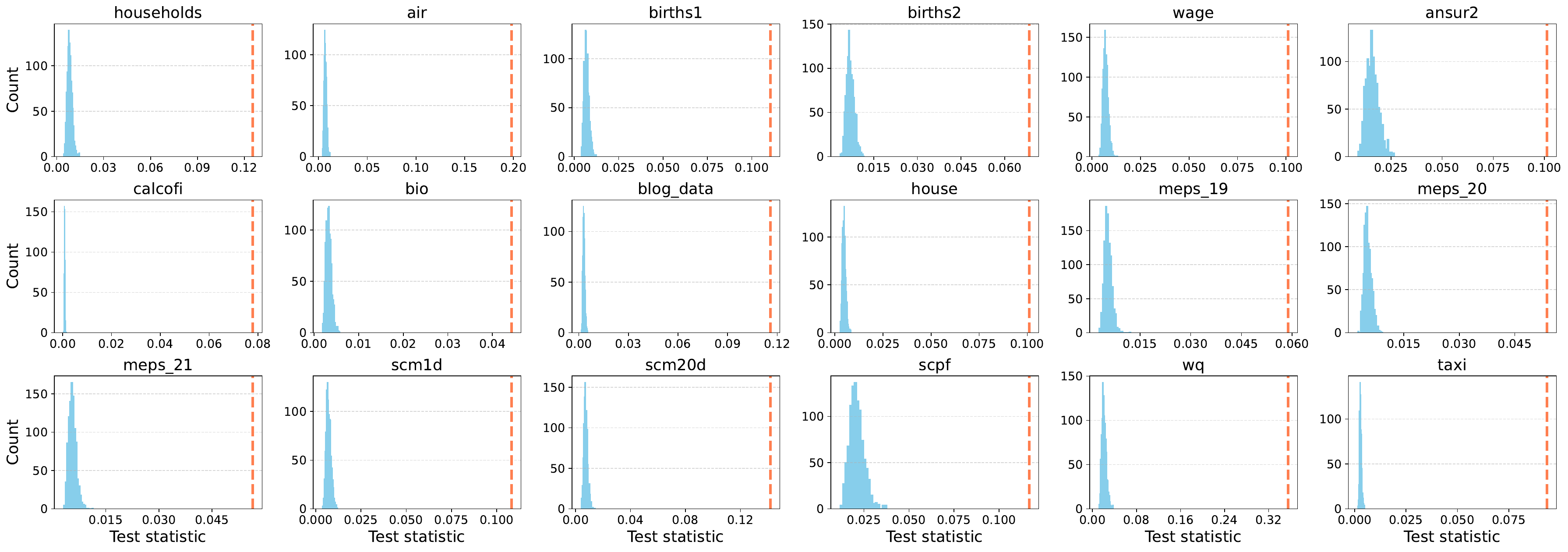}
    \caption{Distributions of the average PCE under the hypothesis of perfect calibration for all datasets, evaluated using the MIX NLL model and the \textbf{scale} prerank.}
    \label{fig:pce_null_vs_real_variance}
\end{figure*}

\begin{figure*}[t]
    \centering
    \includegraphics[width=\textwidth]{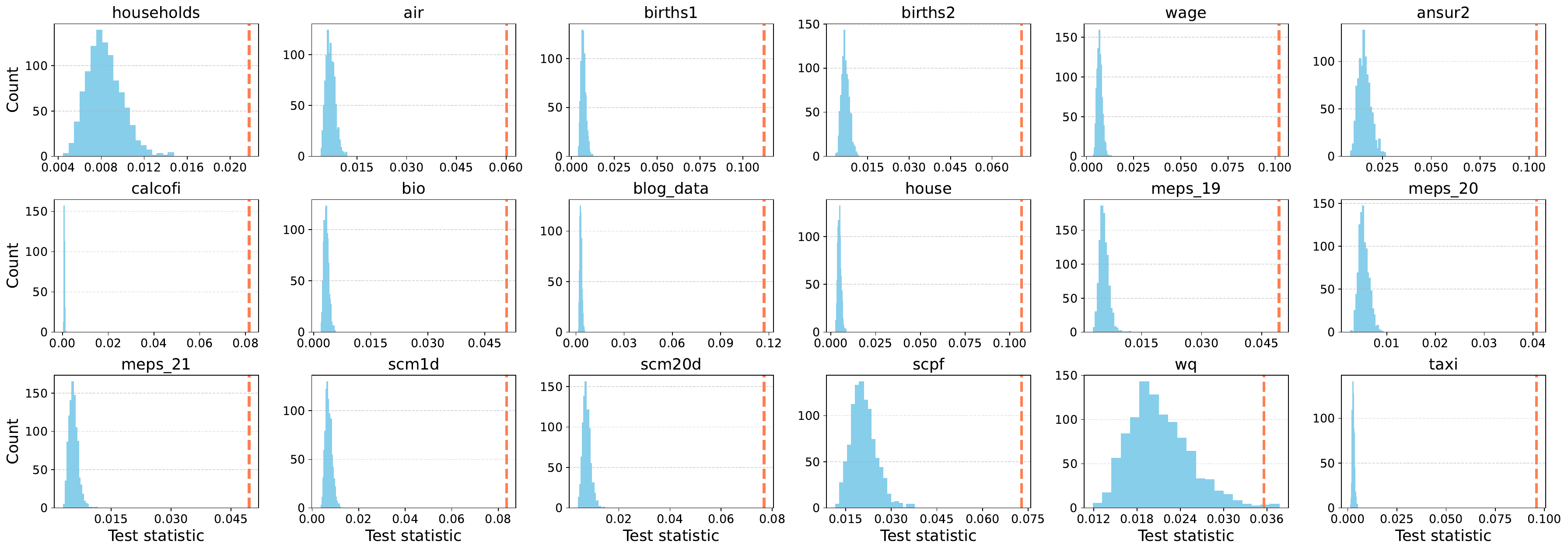}
    \caption{Distributions of the average PCE under the hypothesis of perfect calibration for all datasets, evaluated using the MIX NLL model and the \textbf{dependency} prerank.}
    \label{fig:pce_null_vs_real_dependency}
\end{figure*}

\begin{figure*}[t]
    \centering
    \includegraphics[width=\textwidth]{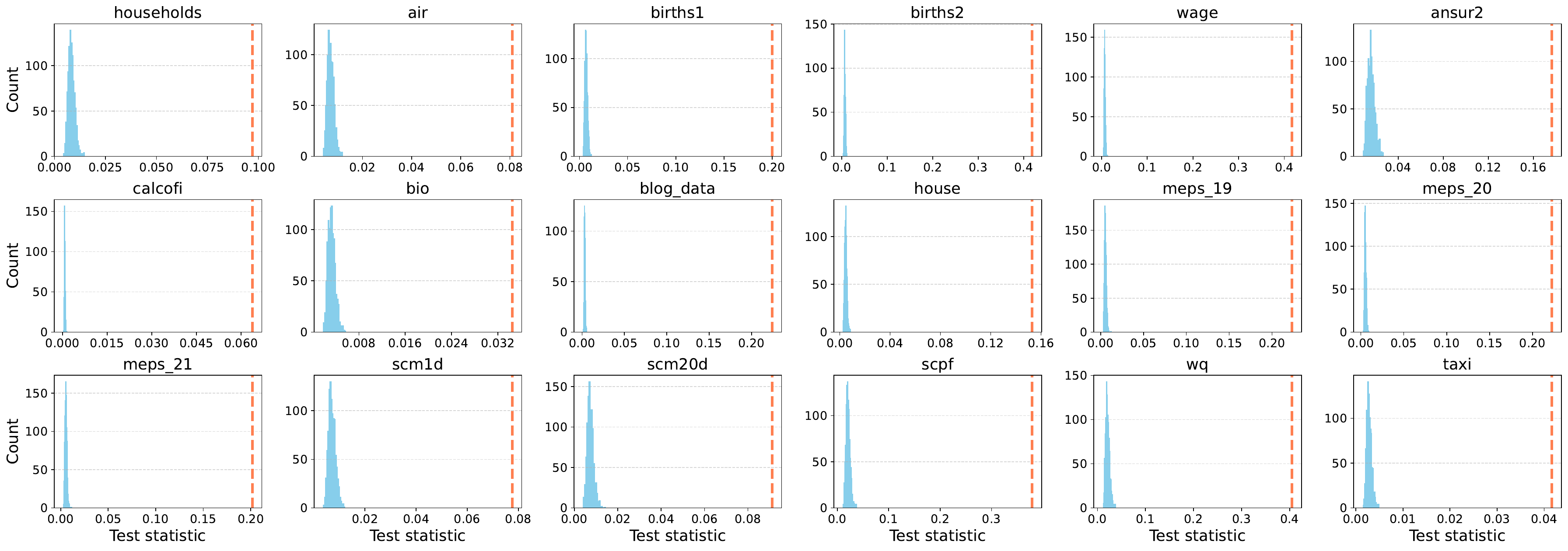}
    \caption{Distributions of the average PCE under the hypothesis of perfect calibration for all datasets, evaluated using the MIX NLL model and the \textbf{HDR} prerank.}
    \label{fig:pce_null_vs_real_density}
\end{figure*}

\begin{figure*}[!htbp]
    \centering
    \includegraphics[width=\textwidth]{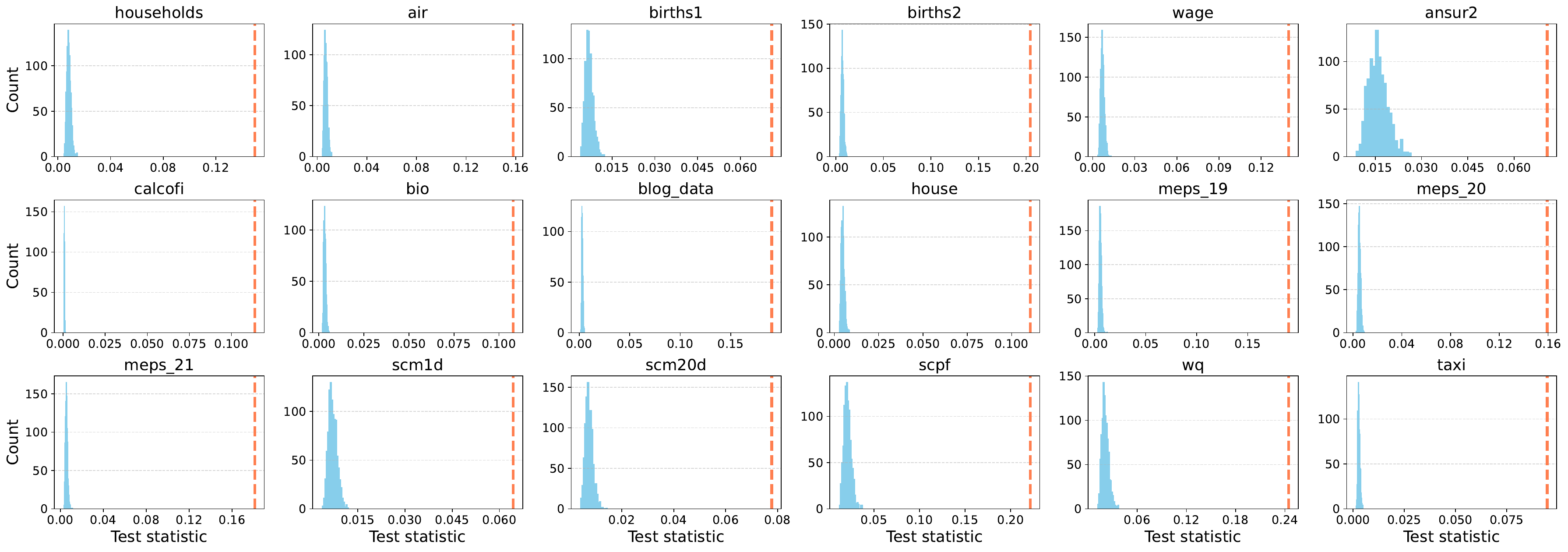}
    \caption{Distributions of the average PCE under the hypothesis of perfect calibration for all datasets, evaluated using the MIX NLL model and the \textbf{Copula} prerank.}
    \label{fig:pce_null_vs_real_cdf}
\end{figure*}

Figures~\ref{fig:pce_null_vs_real_marginal} - \ref{fig:pce_null_vs_real_pca} show histograms of the null distributions of the average (over 5 runs) PCE for each dataset and pre-rank with the corresponding observed averages. In many cases, the observed average PCE lies deep in the right tail of the null distribution; for several datasets, it even exceeds all \(10^4\) simulated averages, demonstrating strong deviations from perfect calibration.

\subsection{D. Hyperparameters}
As described in the main paper, we select the $\lambda$ that minimizes PCE while ensuring that ES does not increase by more than 10\% relative to the reference ES from the best epoch of the model trained with $\lambda = 0$. This tuning is performed separately for each dataset and pre-rank pair. We select $\lambda$ on validation set from $\{0, 0.01, 0.1, 1, 5, 10\}$. Table~\ref{tab:lambda-tuning} shows the selected $\lambda$ for each dataset-pre-rank pair. Notably, the majority of selected values are large, often $\lambda = 10$, suggesting that future work could explore larger values or employ more sophisticated tuning strategies such as Bayesian Optimization.

\begin{table}[!htbp]
\centering
\scriptsize
\begin{tabular}{l|*{7}{c}}
\hline
Datasets & Marginal & Loc. & Scale & Dep. & PCA & HDR & Copula \\
\hline
households & 10.0 & 10.0 & 10.0 & 5.0 & 10.0 & 10.0 & 5.0\\
air & 10.0 & 10.0 & 10.0 & 10.0 & 10.0 & 10.0 & 5.0\\
births1 & 10.0 & 10.0 & 10.0 & 10.0 & 10.0 & 5.0 & 10.0 \\
births2 & 10.0 & 10.0 & 5.0 & 5.0 & 10.0 & 0.01 & 10.0\\
wage & 10.0 & 10.0 & 5.0 & 10.0 & 5.0 & 1.0 & 10.0\\
scm20d & 10.0 & 10.0 & 10.0 & 10.0 & 5.0 & 10.0 & 0.01\\
% rf2 & 5.0 & 1.0 & 10.0 & 10.0 & 0.1 & 10.0 & 0.01\\
% rf1 & 0.1 & 0.0 & 10.0 & 5.0 & 5.0 & 5.0 & 10.0 \\
scm1d & 10.0 & 10.0 & 10.0 & 10.0 & 10.0 & 0.1 & 10.0 \\
% sf2 & 5.0 & 5.0 & 10.0 & 1.0 & 5.0 & 0.01 & 10.0 \\
wq & 5.0 & 10.0 & 10.0 & 5.0 & 5.0 & 5.0 & 10.0 \\
scpf & 5.0 & 10.0 & 10.0 & 0.0 & 5.0 & 1.0 & 10.0 \\
meps21 & 5.0 & 5.0 & 5.0 & 10.0 & 10.0& 10.0 & 5.0 \\
meps19 & 5.0 & 10.0 & 1.0 & 10.0 & 10.0 & 10.0 & 10.0\\
meps20 & 5.0 & 10.0 & 1.0 & 1.0 & 10.0 & 10.0 & 10.0\\
house & 5.0 & 10.0 & 5.0 & 10.0 & 5.0 & 5.0 & 10.0 \\
bio & 5.0 & 10.0 & 10.0 & 10.0 & 5.0 & 10.0 & 5.0\\
blog data & 10.0 & 10.0 & 10.0 & 10.0 & 10.0 & 10.0 & 10.0 \\
calcofi & 10.0 & 5.0 & 10.0 & 10.0 & 10.0 & 10.0 & 5.0\\
ansur2 & 10.0 & 5.0 & 5.0 & 10.0 & 10.0 & 5.0 & 10.0\\
taxi & 10.0 & 10.0 & 10.0 & 5.0 & 10.0 & 5.0 & 10.0 \\
\hline
\end{tabular}
\caption{Values of $\lambda$ after hyperparameter tuning with each regularization and each pre-rank. The baseline used is MIX-NLL.}
\label{tab:lambda-tuning}
\end{table}

\subsection{E. Detailed Results}
\begin{figure*}[t]
\centering
\begin{minipage}{\textwidth}
\centering
\includegraphics[width=\textwidth]{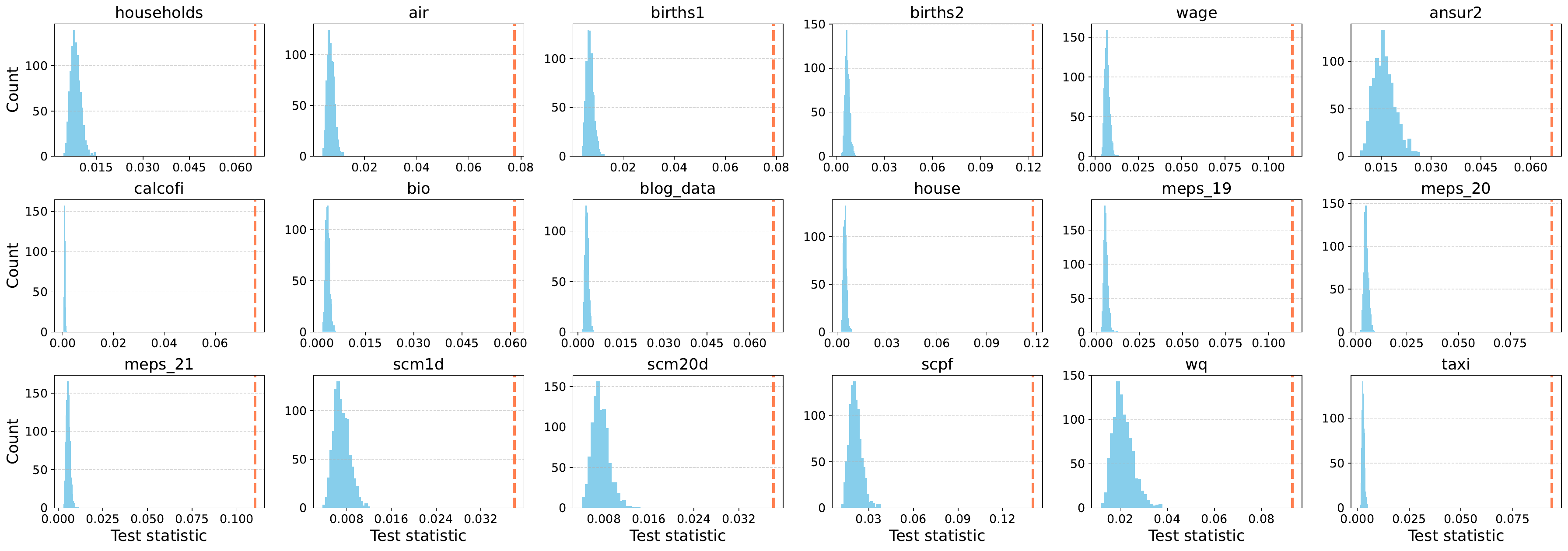}
\captionof{figure}{Distributions of the average PCE under the hypothesis of perfect calibration for all datasets, evaluated using the MIX NLL model and the \textbf{PCA} pre-rank.}
\label{fig:pce_null_vs_real_pca}
\end{minipage}

\vspace{5mm}

\begin{minipage}{\textwidth}
\centering
\scriptsize
\begin{tabular}{l|ccccccc}
\hline
Datasets & Marg. & Loc. & Scale & Dep. & PCA & HDR & Copula \\
\hline
households & 0.095 (0.004) & 0.092 (0.002) & 0.125 (0.006) & 0.022 (0.002) & 0.066 (0.002) & 0.097 (0.007) & 0.149 (0.011) \\
air & 0.099	(0.002) & 0.093	(0.003) & 0.198	(0.005) & 0.060	(0.003) & 0.077	(0.001) & 0.081	(0.007) & 0.158	(0.004) \\
births1 & 0.084	(0.003) & 0.086	(0.003) & 0.110	(0.006) & 0.112 (0.006) & 0.079 (0.002) & 0.200 (0.011) & 0.071 (0.003)\\
births2 & 0.193	(0.008) & 0.050	(0.004) & 0.068	(0.002) & 0.071	(0.002) & 0.123	(0.006) & \textbf{0.418	(0.018)} & 0.204	(0.031) \\
wage & 0.129 (0.001) & 0.082	(0.002) & 0.101	(0.004) & 0.102 (0.004) & 0.114 (0.003) & 0.417 (0.005) & 0.140 (0.003) \\
scm20d & 0.072 (0.001) & 0.077	(0.002) & 0.142	(0.004) & 0.077	(0.007) & 0.038	(0.001) & 0.091	(0.003) & 0.078	(0.001) \\
% rf2 & 0.118	(0.009) & 0.154	(0.014) & 0.136	(0.008) & 0.151	(0.019) & 0.089	(0.004)	& 0.240	(0.050) & 0.137	(0.013)\\
% rf1 & 0.125	(0.008) & 0.162	(0.009) & 0.136	(0.008) & 0.161	(0.010) & 0.091	(0.004) & 0.268	(0.029) & 0.146	(0.009) \\
scm1d & 0.054 (0.003) & 0.058	(0.005) & 0.108	(0.002) & 0.084	(0.009) & 0.038	(0.001) & 0.078	(0.004) & 0.064	(0.004)\\
% sf2 & 0.192	(0.009) & 0.192	(0.017) & 0.128	(0.003) & \textbf{0.266	(0.024)} & \textbf{0.169	(0.009)} & 0.336	(0.006) & 0.174	(0.020) \\
wq & 0.167 (0.005) & 0.104 (0.003) & \textbf{0.355	(0.008)} & 0.036	(0.001) & 0.093 (0.001) & 0.405 (0.007) & \textbf{0.245 (0.001)}\\
scpf & \textbf{0.208 (0.009)} & \textbf{0.210	(0.013)} & 0.117	(0.003) & 0.073	(0.006) & 0.140	(0.005) & 0.379	(0.015) & 0.222	(0.019)\\
meps21 & 0.119 (0.002) & 0.132 (0.002) & 0.056 (0.003) & 0.050 (0.004) & 0.110 (0.005) & 0.202 (0.006) & 0.181 (0.006) \\
meps19 & 0.132 (0.004) & 0.140 (0.005) & 0.059 (0.002) & 0.049 (0.002) & 0.114 (0.004) & 0.223 (0.009) & 0.190 (0.006) \\
meps20 & 0.111 (0.004) & 0.110 (0.007) & 0.054 (0.001) & 0.041 (0.001) & 0.095 (0.004) & 0.223 (0.007) & 0.159 (0.005) \\
house & 0.109 (0.002) & 0.126 (0.002) & 0.101 (0.001) & 0.107 (0.001) & 0.118 (0.002) & 0.153 (0.003) & 0.110 (0.002) \\
bio & 0.057 (0.002) & 0.058 (0.005) & 0.044 (0.002) & 0.051 (0.002) & 0.061 (0.003) & 0.034 (0.002) & 0.108 (0.005) \\
blogdata & 0.138 (0.002) & 0.062 (0.002) & 0.116 (0.004) & 0.117 (0.004) & 0.068 (0.001) & 0.224 (0.003) & 0.191 (0.006) \\
calcofi & 0.075 (0.000) & 0.080 (0.001) & 0.078 (0.000) & 0.082 (0.000) & 0.076 (0.000) & 0.064 (0.001) & 0.114 (0.001) \\
ansur2 & 0.068 (0.004) & 0.070 (0.006) & 0.101 (0.006) & 0.104 (0.006) & 0.066 (0.004) & 0.176 (0.011) & 0.071 (0.005) \\
taxi & 0.082 (0.001) & 0.073 (0.001) & 0.094 (0.003) & 0.096 (0.003) & 0.094 (0.002) & 0.042 (0.001) & 0.095 (0.002) \\
\bottomrule
\end{tabular}
\captionof{table}{Results of real-world experiments using the MIX-NLL model. PCE values are computed using seven pre-rank functions across 18 real datasets and averaged over five runs. Standard errors are shown in parentheses.}
\label{tab:pce-before}
\end{minipage}
\end{figure*}

\noindent \textbf{Pre-rank Calibration.} Table \ref{tab:real-pce-after} reports the exact PCE values averaged over five runs for each pre-rank on which the MIX-NLL+PCE-KDE model was trained using the optimal $\lambda$. For comparison, we also include the PCE values computed with respect to each pre-rank for the baseline MIX-NLL model trained without regularization (see Table \ref{tab:pce-before}). Note that although the baseline model was not trained with respect to any specific pre-rank, we still evaluate its performance on each pre-rank to highlight the benefit of regularization.

\noindent \textbf{Marginal and Pre-rank Calibration.}
Figures~\ref{fig:prerank-households}–\ref{fig:prerank-blog-data} show reliability plots for the majority of real benchmark datasets using the MIX-NLL+PCE-KDE model trained with each of the pre-ranks. We display only a representative subset of datasets due to the similarity of plots. In each figure, the top row shows calibration curves with respect to the pre-rank used during training, while the bottom row shows marginal calibration plots from the same models.

We observe that in many cases, the top-row plots exhibit strong alignment between the empirical CDF and the quantile levels $\alpha$, indicating effective calibration with respect to the training pre-rank. However, the bottom-row plots reveal that strong calibration with respect to the pre-rank does not necessarily lead to better marginal calibration.

In contrast, Figures~\ref{fig:marg-prerank-households}–\ref{fig:marg-prerank-blog} present the reliability diagrams for the same datasets using the MIX-NLL+PCE-KDE model trained with the marginal+pre-rank objective. These plots demonstrate that jointly enforcing calibration with respect to both the marginal and the pre-ranks consistently results in strong calibration across both aspects.

\noindent \textbf{PCA and Pre-rank Calibration.}
In the main paper, we reported calibration results on the \texttt{scm1d} dataset with 16 target dimensions, comparing four training variants: None (no regularization), pre-rank regularization, marginal+pre-rank, and PCA+pre-rank. We extend these observations by presenting additional PCE scores for datasets with target dimension $D \geq 4$. \\
These results confirm that the PCA+pre-rank approach consistently maintains competitive calibration performance using only a small number of informative projections, offering a practical solution for higher-dimensional multivariate regression tasks. In general, PCA offers a good trade-off between calibration quality and computational cost: it often achieves marginal and pre-rank-specific calibration performance close to that of the marginal+pre-rank method, while being significantly more efficient to compute. The full set of results is reported in Tables \ref{tab:scm20d-prerank} - \ref{tab:households-prerank}.

\clearpage

\begin{table*}[t]
\centering
\begin{minipage}[t]{0.48\textwidth}
\begin{table}[H]
\centering
\small
\resizebox{\textwidth}{!}{
\begin{tabular}{@{}l|ccccccc@{}}
\hline
Method & Marg & Loc & Scale & Dep & PCA & HDR & Cop \\
\hline
None & \textbf{0.072} & \textbf{0.077} & \textbf{0.142} & \textbf{0.077} & \textbf{0.038} & \textbf{0.091} & \textbf{0.078} \\
\hline
Marg & \textbf{0.033} & 0.035 & 0.094 & 0.042 & 0.033 & 0.085 & 0.037 \\
Loc & \textbf{0.036} & \textbf{0.025} & 0.109 & 0.053 & 0.034 & 0.086 & 0.046 \\
Scale & \textbf{0.053} & 0.071 & \textbf{0.042} & 0.054 & 0.039 & 0.107 & 0.055 \\
Dep & \textbf{0.072} & 0.081 & 0.117 & \textbf{0.025} & 0.038 & 0.08 & 0.072 \\
PCA & \textbf{0.048} & 0.045 & 0.114 & 0.058 & \textbf{0.033} & 0.086 & 0.054 \\
HDR & \textbf{0.082} & 0.086 & 0.167 & 0.096 & 0.039 & \textbf{0.061} & 0.086 \\
Cop & \textbf{0.042} & 0.051 & 0.125 & 0.048 & 0.033 & 0.084 & \textbf{0.031} \\
\hline
Marg+loc & \textbf{0.029} & \textbf{0.024} & 0.101 & 0.045 & 0.032 & 0.084 & 0.038 \\
Marg+scale & \textbf{0.027} & 0.033 & \textbf{0.041} & 0.045 & 0.032 & 0.077 & 0.035 \\
Marg+dep & \textbf{0.037} & 0.041 & 0.085 & \textbf{0.026} & 0.033 & 0.081 & 0.038 \\
Marg+HDR & \textbf{0.041} & 0.044 & 0.123 & 0.044 & 0.032 & \textbf{0.088} & 0.042 \\
Marg+Cop & \textbf{0.031} & 0.035 & 0.096 & 0.038 & 0.032 & 0.086 & \textbf{0.028} \\
\hline
PCA+loc & \textbf{0.034} & \textbf{0.023} & 0.102 & 0.035 & 0.031 & 0.087 & 0.045 \\
PCA+scale & \textbf{0.034} & 0.029 & \textbf{0.045} & 0.041 & 0.03 & 0.083 & 0.037 \\
PCA+dep & \textbf{0.045} & 0.039 & 0.097 & \textbf{0.024} & 0.033 & 0.083 & 0.05 \\
PCA+HDR & \textbf{0.05} & 0.043 & 0.123 & 0.044 & 0.034 & \textbf{0.089} & 0.054 \\
PCA+Cop & \textbf{0.032} & 0.028 & 0.09 & 0.044 & 0.032 & 0.082 & \textbf{0.028} \\
\hline
\end{tabular}
}
\caption{PCE values for the \textbf{scm20d} dataset.}
\label{tab:scm20d-prerank}
\end{table}

\vspace{-3mm}

\begin{table}[H]
\centering
\small
\resizebox{\textwidth}{!}{
\begin{tabular}{@{}l|ccccccc@{}}
\hline
Method & Marg & Loc & Scale & Dep & PCA & HDR & Cop \\
\hline
None & \textbf{0.167} & \textbf{0.104} & \textbf{0.355} & \textbf{0.036} & \textbf{0.093} & \textbf{0.405} & \textbf{0.245} \\
\hline
Marg & \textbf{0.154} & 0.117 & 0.34 & 0.027 & 0.088 & 0.379 & 0.245 \\
Loc & \textbf{0.152} & \textbf{0.074} & 0.336 & 0.026 & 0.086 & 0.359 & 0.236 \\
Scale & \textbf{0.156} & 0.109 & \textbf{0.31} & 0.027 & 0.084 & 0.365 & 0.246 \\
Dep & \textbf{0.167} & 0.116 & 0.346 & \textbf{0.026} & 0.094 & 0.4 & 0.245 \\
PCA & \textbf{0.163} & 0.113 & 0.344 & 0.027 & \textbf{0.092} & 0.387 & 0.245 \\
HDR & \textbf{0.159} & 0.109 & 0.34 & 0.025 & 0.09 & \textbf{0.375} & 0.244 \\
Cop & \textbf{0.162} & 0.108 & 0.339 & 0.033 & 0.09 & 0.381 & \textbf{0.243} \\
\hline
Marg+loc & \textbf{0.151} & \textbf{0.098} & 0.338 & 0.027 & 0.085 & 0.373 & 0.24 \\
Marg+scale & \textbf{0.135} & 0.129 & \textbf{0.291} & 0.034 & 0.082 & 0.339 & 0.247 \\
Marg+dep & \textbf{0.145} & 0.128 & 0.332 & \textbf{0.028} & 0.088 & 0.365 & 0.244 \\
Marg+HDR & \textbf{0.14} & 0.137 & 0.318 & 0.027 & 0.087 & \textbf{0.349} & 0.243 \\
Marg+Cop & \textbf{0.15} & 0.133 & 0.34 & 0.026 & 0.092 & 0.382 & \textbf{0.245} \\
\hline
PCA+loc & \textbf{0.155} & \textbf{0.088} & 0.33 & 0.022 & 0.088 & 0.366 & 0.237 \\
PCA+scale & \textbf{0.146} & 0.103 & \textbf{0.294} & 0.031 & 0.08 & 0.334 & 0.246 \\
PCA+dep & \textbf{0.161} & 0.106 & 0.335 & \textbf{0.027} & 0.086 & 0.384 & 0.245 \\
PCA+HDR & \textbf{0.146} & 0.109 & 0.319 & 0.028 & 0.081 & \textbf{0.35} & 0.243 \\
PCA+Cop & \textbf{0.161} & 0.109 & 0.336 & 0.032 & 0.088 & 0.386 & \textbf{0.244} \\
\hline
\end{tabular}
}
\caption{PCE values for the \textbf{wq} dataset.}
\label{tab:wq-prerank}
\end{table}

\vspace{-3mm}

\begin{table}[H]
\centering
\small
\resizebox{\textwidth}{!}{
\begin{tabular}{@{}l|ccccccc@{}}
\hline
Method & Marg & Loc & Scale & Dep & PCA & HDR & Cop \\
\hline
None & \textbf{0.099} & \textbf{0.093} & \textbf{0.198} & \textbf{0.060} & \textbf{0.077} & \textbf{0.081} & \textbf{0.158} \\
\hline
Marg & \textbf{0.041} & 0.043 & 0.114 & 0.046 & 0.045 & 0.037 & 0.045 \\
Loc & \textbf{0.059} & \textbf{0.027} & 0.158 & 0.061 & 0.054 & 0.045 & 0.091 \\
Scale & \textbf{0.114} & 0.129 & \textbf{0.024} & 0.069 & 0.075 & 0.136 & 0.14 \\
Dep & \textbf{0.094} & 0.091 & 0.2 & \textbf{0.027} & 0.072 & 0.068 & 0.142 \\
PCA & \textbf{0.047} & 0.033 & 0.074 & 0.06 & \textbf{0.039} & 0.065 & 0.057 \\
HDR & \textbf{0.092} & 0.091 & 0.164 & 0.063 & 0.063 & \textbf{0.033} & 0.137 \\
Cop & \textbf{0.062} & 0.098 & 0.2 & 0.057 & 0.069 & 0.067 & \textbf{0.033} \\
\hline
Marg+loc & \textbf{0.039} & \textbf{0.026} & 0.117 & 0.06 & 0.043 & 0.044 & 0.071 \\
Marg+scale & \textbf{0.037} & 0.043 & \textbf{0.024} & 0.05 & 0.039 & 0.071 & 0.059 \\
Marg+dep & \textbf{0.04} & 0.043 & 0.105 & \textbf{0.024} & 0.043 & 0.042 & 0.049 \\
Marg+HDR & \textbf{0.042} & 0.047 & 0.124 & 0.046 & 0.046 & \textbf{0.039} & 0.049 \\
Marg+Cop & \textbf{0.041} & 0.052 & 0.129 & 0.044 & 0.049 & 0.036 & \textbf{0.032} \\
\hline
PCA+loc & \textbf{0.051} & \textbf{0.026} & 0.09 & 0.046 & 0.046 & 0.05 & 0.08 \\
PCA+scale & \textbf{0.053} & 0.046 & \textbf{0.024} & 0.039 & 0.046 & 0.087 & 0.075 \\
PCA+dep & \textbf{0.052} & 0.037 & 0.091 & \textbf{0.025} & 0.047 & 0.047 & 0.079 \\
PCA+HDR & \textbf{0.049} & 0.038 & 0.109 & 0.049 & 0.046 & \textbf{0.036} & 0.069 \\
PCA+Cop & \textbf{0.043} & 0.038 & 0.09 & 0.049 & 0.044 & 0.037 & \textbf{0.037} \\
\hline
\end{tabular}
}
\caption{PCE values for the \textbf{air} dataset.}
\label{tab:air-prerank}
\end{table}
\end{minipage}\hfill
\begin{minipage}[t]{0.48\textwidth}
\begin{table}[H]
\centering
\small
\resizebox{\textwidth}{!}{
\begin{tabular}{@{}l|ccccccc@{}}
\hline
Method & Marg & Loc & Scale & Dep & PCA & HDR & Cop \\
\hline
None & \textbf{0.193} & \textbf{0.050} & \textbf{0.068} & \textbf{0.071} & \textbf{0.123} & \textbf{0.418} & \textbf{0.204} \\
\hline
Marg & \textbf{0.05} & 0.034 & 0.062 & 0.037 & 0.072 & 0.345 & 0.078 \\
Loc & \textbf{0.18} & \textbf{0.03} & 0.051 & 0.061 & 0.117 & 0.344 & 0.146 \\
Scale & \textbf{0.192} & 0.039 & \textbf{0.047} & 0.073 & 0.117 & 0.352 & 0.187 \\
Dep & \textbf{0.175} & 0.061 & 0.079 & \textbf{0.031} & 0.129 & 0.388 & 0.178 \\
PCA & \textbf{0.093} & 0.033 & 0.05 & 0.053 & \textbf{0.062} & 0.327 & 0.083 \\
HDR & \textbf{0.21} & 0.054 & 0.068 & 0.076 & 0.145 & \textbf{0.417} & 0.248 \\
Cop & \textbf{0.174} & 0.037 & 0.056 & 0.077 & 0.124 & 0.368 & \textbf{0.043} \\
\hline
Marg+loc & \textbf{0.033} & \textbf{0.03} & 0.061 & 0.047 & 0.052 & 0.355 & 0.047 \\
Marg+scale & \textbf{0.033} & 0.03 & \textbf{0.044} & 0.043 & 0.041 & 0.335 & 0.048 \\
Marg+dep & \textbf{0.032} & 0.036 & 0.06 & \textbf{0.03} & 0.053 & 0.347 & 0.057 \\
Marg+HDR & \textbf{0.043} & 0.05 & 0.097 & 0.036 & 0.044 & \textbf{0.04} & 0.054 \\
Marg+Cop & \textbf{0.035} & 0.032 & 0.05 & 0.045 & 0.04 & 0.32 & \textbf{0.036} \\
\hline
PCA+loc & \textbf{0.123} & \textbf{0.028} & 0.047 & 0.053 & 0.071 & 0.363 & 0.067 \\
PCA+scale & \textbf{0.118} & 0.031 & \textbf{0.043} & 0.057 & 0.067 & 0.371 & 0.065 \\
PCA+dep & \textbf{0.12} & 0.039 & 0.051 & \textbf{0.03} & 0.076 & 0.375 & 0.095 \\
PCA+HDR & \textbf{0.085} & 0.044 & 0.107 & 0.033 & 0.068 & \textbf{0.048} & 0.052 \\
PCA+Cop & \textbf{0.1} & 0.03 & 0.054 & 0.063 & 0.074 & 0.344 & \textbf{0.035} \\
\hline
\end{tabular}
}
\caption{PCE values for the \textbf{births2} dataset.}
\label{tab:births2-prerank}
\end{table}

\vspace{-3mm}

\begin{table}[H]
\centering
\small
\resizebox{\textwidth}{!}{
\begin{tabular}{@{}l|ccccccc@{}}
\hline
Method & Marg & Loc & Scale & Dep & PCA & HDR & Cop \\
\hline
None & \textbf{0.095} & \textbf{0.092} & \textbf{0.125} & \textbf{0.022} & \textbf{0.066} & \textbf{0.097} & \textbf{0.149} \\
\hline
Marg & \textbf{0.028} & 0.029 & 0.057 & 0.019 & 0.029 & 0.032 & 0.028 \\
Loc & \textbf{0.05} & \textbf{0.023} & 0.08 & 0.021 & 0.039 & 0.039 & 0.088 \\
Scale & \textbf{0.102} & 0.131 & \textbf{0.021} & 0.028 & 0.067 & 0.073 & 0.118 \\
Dep & \textbf{0.098} & 0.097 & 0.127 & \textbf{0.019} & 0.067 & 0.073 & 0.154 \\
PCA & \textbf{0.049} & 0.025 & 0.043 & 0.027 & \textbf{0.025} & 0.038 & 0.073 \\
HDR & \textbf{0.097} & 0.102 & 0.076 & 0.025 & 0.059 & \textbf{0.029} & 0.141 \\
Cop & \textbf{0.046} & 0.062 & 0.124 & 0.018 & 0.055 & 0.066 & \textbf{0.031} \\
\hline
Marg+loc & \textbf{0.029} & \textbf{0.017} & 0.048 & 0.024 & 0.03 & 0.033 & 0.054 \\
Marg+scale & \textbf{0.028} & 0.034 & \textbf{0.021} & 0.034 & 0.026 & 0.038 & 0.032 \\
Marg+dep & \textbf{0.029} & 0.03 & 0.062 & \textbf{0.02} & 0.031 & 0.031 & 0.029 \\
Marg+HDR & \textbf{0.032} & 0.029 & 0.061 & 0.024 & 0.033 & \textbf{0.034} & 0.04 \\
Marg+Cop & \textbf{0.029} & 0.035 & 0.049 & 0.025 & 0.033 & 0.027 & \textbf{0.028} \\
\hline
PCA+loc & \textbf{0.043} & \textbf{0.02} & 0.062 & 0.031 & 0.032 & 0.033 & 0.067 \\
PCA+scale & \textbf{0.043} & 0.029 & \textbf{0.021} & 0.043 & 0.031 & 0.031 & 0.046 \\
PCA+dep & \textbf{0.045} & 0.026 & 0.074 & \textbf{0.02} & 0.034 & 0.039 & 0.069 \\
PCA+HDR & \textbf{0.043} & 0.025 & 0.059 & 0.023 & 0.031 & \textbf{0.035} & 0.059 \\
PCA+Cop & \textbf{0.037} & 0.025 & 0.032 & 0.022 & 0.029 & 0.036 & \textbf{0.03} \\
\hline
\end{tabular}
}
\caption{PCE values for the \textbf{households} dataset.}
\label{tab:households-prerank}
\end{table}
\end{minipage}
\end{table*}

%this one

%this one

%this one

%this one

%this one

%this one

% \begin{table*}[t]
% \centering
% \scriptsize
% \begin{tabular}{l|ccccccc}
% \hline
% Datasets & Marg. & Loc. & Scale & Dep. & PCA & HDR & Copula \\
% \hline

% \end{tabular}
% \caption{Results of real-world experiments using the MIX-NLL model. PCE values are computed using seven pre-rank functions across 18 real datasets and averaged over five runs. Standard errors are shown in parentheses.}
% \label{tab:pce-before}
% \end{table*}

\clearpage

\begin{table*}[t]
\centering
\scriptsize
\begin{tabular}{l|ccccccc}
\hline
Datasets & Marg. & Loc. & Scale & Dep. & PCA & HDR & Copula \\
\hline
households & 0.030 (0.001) & 0.024 (0.002) & 0.021 (0.002) & 0.021 (0.003) & 0.026 (0.001) & 0.039 (0.003) & 0.031 (0.001) \\
air & 0.040 (0.001) & 0.026 (0.002) & 0.027 (0.001) & 0.027 (0.002) & 0.039 (0.002) & 0.045 (0.005) & 0.029 (0.001) \\
births1 & 0.028 (0.001) & 0.027 (0.002) & 0.031 (0.002) & 0.027 (0.002) & 0.034 (0.000) & 0.034 (0.003) & 0.027 (0.001) \\
births2 & 0.031 (0.002) & 0.029 (0.002) & 0.045 (0.002) & 0.032 (0.001) & 0.052 (0.006) & \textbf{0.428 (0.002)} & 0.033 (0.002) \\
wage & 0.051 (0.020) & 0.044 (0.013) & 0.025 (0.001) & 0.024 (0.002) & 0.052 (0.008) & 0.364 (0.012) & 0.095 (0.015) \\
scm20d & 0.033 (0.002) & 0.025 (0.001) & 0.038 (0.003) & 0.022 (0.002) & 0.033 (0.001) & 0.093 (0.003) & 0.029 (0.001) \\
% \elnura{rf2} & & & & & & & \\
% % & 0.088 (0.022) & 0.154 (0.012) & 0.045 (0.018) & 0.047 (0.021) & 0.075 (0.003) & 0.066 (0.034) & 0.118 (0.019)\\
% \elnura{rf1} & & & & & & & \\
scm1d & 0.025 (0.001) & 0.024 (0.002) & 0.036 (0.003) & 0.020 (0.002) & 0.035 (0.001) & 0.090 (0.015) & 0.036 (0.007) \\
% \elnura{sf2} & & & & & & & \\
wq & \textbf{0.154 (0.007)} & \textbf{0.076 (0.010)} & \textbf{0.311 (0.018)} & 0.028 (0.003) & \textbf{0.091 (0.004)} & 0.376 (0.026) & \textbf{0.244 (0.001)} \\
scpf & 0.039 (0.002) & 0.026 (0.003) & 0.041 (0.005) & \textbf{0.081 (0.005)} & 0.063 (0.004) & 0.299 (0.010) & 0.032 (0.006) \\
meps21 & 0.026 (0.001) & 0.025 (0.001) & 0.031 (0.001) & 0.024 (0.001) & 0.025 (0.001) & 0.032 (0.001) & 0.023 (0.001) \\
meps19 & 0.026 (0.002) & 0.023 (0.001) & 0.050 (0.002) & 0.025 (0.002) & 0.024 (0.001) & 0.031 (0.001) & 0.022 (0.001) \\
meps20 & 0.024 (0.001) & 0.023 (0.001) & 0.049 (0.001) & 0.033 (0.001) & 0.024 (0.000) & 0.034 (0.005) & 0.024 (0.001) \\
house & 0.027 (0.003) & 0.020 (0.001) & 0.025 (0.003) & 0.020 (0.001) & 0.033 (0.004) & 0.028 (0.002) & 0.021 (0.000) \\
bio & 0.021 (0.001) & 0.020 (0.001) & 0.021 (0.001) & 0.021 (0.001) & 0.021 (0.001) & 0.021 (0.001) & 0.021 (0.001) \\
blogdata & 0.023 (0.001) & 0.024 (0.000) & 0.023 (0.001) & 0.023 (0.001) & 0.025 (0.000) & 0.030 (0.001) & 0.024 (0.001) \\
calcofi & 0.020 (0.000) & 0.021 (0.000) & 0.020 (0.000) & 0.020 (0.000) & 0.021 (0.000) & 0.020 (0.000) & 0.020 (0.000) \\
ansur2 & 0.032 (0.004) & 0.040 (0.009) & 0.031 (0.004) & 0.025 (0.003) & 0.038 (0.003) & 0.040 (0.015) & 0.039 (0.009) \\
taxi & 0.021 (0.001) & 0.022 (0.001) & 0.025 (0.000) & 0.023 (0.001) & 0.022 (0.001) & 0.022 (0.000) & 0.022 (0.001) \\
\bottomrule
\end{tabular}
\caption{PCE values after applying PCE-KDE regularization with MIX-NLL using the optimal $\lambda$. Results are reported for seven pre-rank functions across 18 real datasets, averaged over five runs. Standard errors are shown in parentheses.}
\label{tab:real-pce-after}
\end{table*}

%this one
\begin{figure*}[t]
    \centering
    \includegraphics[width=\textwidth]{Figures/rel-plots/rel_grid_prerank_households.pdf}
    \caption{Reliability plots on \texttt{households} dataset using MIX-NLL+PCE-KDE on pre-rank. Top row: calibration curves with respect to: marginal, location, scale, dependency, PCA, HDR, and Copula. Bottom row: corresponding marginal calibration curves.}
    \label{fig:prerank-households}
\end{figure*}

%this one
\begin{figure*}[!htbp]
    \centering
    \includegraphics[width=\textwidth]{Figures/rel-plots/rel_grid_prerank_air.pdf}
    \caption{Reliability plots on \texttt{air} dataset using MIX-NLL+PCE-KDE on pre-rank. Top row: calibration curves with respect to: marginal, location, scale, dependency, PCA, HDR, and Copula. Bottom row: corresponding marginal calibration curves.}
    \label{fig:prerank-air}
\end{figure*}

% \begin{figure*}[!htbp]
%     \centering
%     \includegraphics[width=\textwidth]{Figures/rel-plots/rel_grid_prerank_births1.png}
%     \caption{}
%     \label{fig:}
% \end{figure*}

%this one
\begin{figure*}[!htbp]
    \centering
    \includegraphics[width=\textwidth]{Figures/rel-plots/rel_grid_prerank_births2.pdf}
    \caption{Reliability plots on \texttt{births2} dataset using MIX-NLL+PCE-KDE on pre-rank. Top row: calibration curves with respect to: marginal, location, scale, dependency, PCA, HDR, and Copula. Bottom row: corresponding marginal calibration curves.}
    \label{fig:prerank-births2}
\end{figure*}

\begin{figure*}[!htbp]
    \centering
    \includegraphics[width=\textwidth]{Figures/rel-plots/rel_grid_prerank_meps19.pdf}
    \caption{Reliability plots on \texttt{meps19} dataset using MIX-NLL+PCE-KDE on pre-rank. Top row: calibration curves with respect to: marginal, location, scale, dependency, PCA, HDR, and Copula. Bottom row: corresponding marginal calibration curves.}
    \label{fig:prerank-meps19}
\end{figure*}

% \begin{figure*}[!htbp]
%     \centering
%     \includegraphics[width=\textwidth]{Figures/rel-plots/rel_grid_prerank_meps20.png}
%     \caption{}
%     \label{fig:}
% \end{figure*}

% \begin{figure*}[!htbp]
%     \centering
%     \includegraphics[width=\textwidth]{Figures/rel-plots/rel_grid_prerank_house.png}
%     \caption{Reliability plots on \texttt{house} dataset using MIX-NLL+PCE-KDE on pre-rank. Top row: calibration curves with respect to: marginal, location, scale, dependency, PCA, HDR, and Copula. Bottom row: corresponding marginal calibration curves.}
%     \label{fig:prerank-house}
% \end{figure*}

% \begin{figure*}[!htbp]
%     \centering
%     \includegraphics[width=\textwidth]{Figures/rel-plots/rel_grid_prerank_bio.png}
%     \caption{}
%     \label{fig:}
% \end{figure*}

%this one
\begin{figure*}[!htbp]
    \centering
    \includegraphics[width=\textwidth]{Figures/rel-plots/rel_grid_prerank_blog_data.pdf}
    \caption{Reliability plots on \texttt{blog data} dataset using MIX-NLL+PCE-KDE on pre-rank. Top row: calibration curves with respect to: marginal, location, scale, dependency, PCA, HDR, and Copula. Bottom row: corresponding marginal calibration curves.}
    \label{fig:prerank-blog-data}
\end{figure*}

% \begin{figure*}[!htbp]
%     \centering
%     \includegraphics[width=\textwidth]{Figures/rel-plots/rel_grid_prerank_ansur2.png}
%     \caption{}
%     \label{fig:}
% \end{figure*}

% \begin{figure*}[!htbp]
%     \centering
%     \includegraphics[width=\textwidth]{Figures/rel-plots/rel_grid_prerank_taxi.png}
%     \caption{Reliability plots on \texttt{taxi} dataset using MIX-NLL+PCE-KDE on pre-rank. Top row: calibration curves with respect to: marginal, location, scale, dependency, PCA, HDR, and Copula. Bottom row: corresponding marginal calibration curves.}
%     \label{fig:prerank-taxi}
% \end{figure*}

%this one
\begin{figure*}[!htbp]
    \centering
    \includegraphics[width=\textwidth]{Figures/rel-plots/rel_grid_marg_prerank_households.pdf}
    \caption{Reliability plots on \texttt{households} dataset using MIX-NLL+PCE-KDE on marginal+pre-rank. Top row: calibration curves with respect to: marginal, location, scale, dependency, PCA, HDR, and Copula. Bottom row: corresponding marginal calibration curves.}
    \label{fig:marg-prerank-households}
\end{figure*}

%this one
\begin{figure*}[!htbp]
    \centering
    \includegraphics[width=\textwidth]{Figures/rel-plots/rel_grid_marg_prerank_air.pdf}
    \caption{Reliability plots on \texttt{air} dataset using MIX-NLL+PCE-KDE on marginal+pre-rank. Top row: calibration curves with respect to: marginal, location, scale, dependency, PCA, HDR, and Copula. Bottom row: corresponding marginal calibration curves.}
    % \label{fig:}
\end{figure*}

% \begin{figure*}[!htbp]
%     \centering
%     \includegraphics[width=\textwidth]{Figures/rel-plots/rel_grid_marg_prerank_births1.png}
%     \caption{}
%     \label{fig:}
% \end{figure*}

%this one
\begin{figure*}[!htbp]
    \centering
    \includegraphics[width=\textwidth]{Figures/rel-plots/rel_grid_marg_prerank_births2.pdf}
    \caption{Reliability plots on \texttt{births2} dataset using MIX-NLL+PCE-KDE on marginal+pre-rank. Top row: calibration curves with respect to: marginal, location, scale, dependency, PCA, HDR, and Copula. Bottom row: corresponding marginal calibration curves.}
    % \label{fig:}
\end{figure*}

% \begin{figure*}[!htbp]
%     \centering
%     \includegraphics[width=\textwidth]{Figures/rel-plots/rel_grid_marg_prerank_wage.png}
%     \caption{Reliability plots on \texttt{wage} dataset using MIX-NLL+PCE-KDE on marginal+pre-rank. Top row: calibration curves with respect to: marginal, location, scale, dependency, PCA, HDR, and Copula. Bottom row: corresponding marginal calibration curves.}
%     % \label{fig:}
% \end{figure*}

% \begin{figure*}[!htbp]
%     \centering
%     \includegraphics[width=\textwidth]{Figures/rel-plots/rel_grid_marg_prerank_scm20d.png}
%     \caption{}
%     \label{fig:}
% \end{figure*}

% \begin{figure*}[!htbp]
%     \centering
%     \includegraphics[width=\textwidth]{Figures/rel-plots/rel_grid_marg_prerank_scm1d.png}
%     \caption{}
%     \label{fig:}
% \end{figure*}

% \begin{figure*}[!htbp]
%     \centering
%     \includegraphics[width=\textwidth]{Figures/rel-plots/rel_grid_marg_prerank_wq.png}
%     \caption{}
%     \label{fig:}
% \end{figure*}

% \begin{figure*}[!htbp]
%     \centering
%     \includegraphics[width=\textwidth]{Figures/rel-plots/rel_grid_marg_prerank_scpf.png}
%     \caption{Reliability plots on \texttt{scpf} dataset using MIX-NLL+PCE-KDE on marginal+pre-rank. Top row: calibration curves with respect to: marginal, location, scale, dependency, PCA, HDR, and Copula. Bottom row: corresponding marginal calibration curves.}
%     % \label{fig:}
% \end{figure*}

%this one
\begin{figure*}[p]
    \vspace*{-7cm}
    \centering
    \begin{minipage}{\textwidth}
        \includegraphics[width=\textwidth]{Figures/rel-plots/rel_grid_marg_prerank_meps_19.pdf}
        \caption{Reliability plots on \texttt{meps19} dataset using MIX-NLL+PCE-KDE on marginal+pre-rank. Top row: calibration curves with respect to: marginal, location, scale, dependency, PCA, HDR, and Copula. Bottom row: corresponding marginal calibration curves.}
    \end{minipage}
    
    \vspace{3em}
    
    \begin{minipage}{\textwidth}
        \includegraphics[width=\textwidth]{Figures/rel-plots/rel_grid_marg_prerank_blog_data.pdf}
        \caption{Reliability plots on \texttt{blog data} dataset using MIX-NLL+PCE-KDE on marginal+pre-rank. Top row: calibration curves with respect to: marginal, location, scale, dependency, PCA, HDR, and Copula. Bottom row: corresponding marginal calibration curves.}
        \label{fig:marg-prerank-blog}
    \end{minipage}
\end{figure*}

\end{document}